\documentclass[conference]{IEEEtran}
\IEEEoverridecommandlockouts
\usepackage{caption}
\usepackage{subcaption}
\usepackage{cite}
\usepackage{amsmath,amssymb,amsfonts}
\usepackage{algorithmic}
\usepackage{graphicx}
\usepackage{textcomp}
\usepackage{xcolor}
\usepackage{url}

\usepackage{booktabs} 
\usepackage{tabularx}

\usepackage{float} 
\usepackage{adjustbox}
\usepackage{url}

\definecolor{bgcolor}{rgb}{0.95,0.95,0.95}

\usepackage{multirow}

\def\BibTeX{{\rm B\kern-.05em{\sc i\kern-.025em b}\kern-.08em
    T\kern-.1667em\lower.7ex\hbox{E}\kern-.125emX}}

\usepackage{listings}
\usepackage{array,booktabs}
\newcommand {\otoprule}{\midrule [\heavyrulewidth]}
\newcolumntype {+}{ >{\global\let\currentrowstyle\relax}}
\newcolumntype {^}{ >{\currentrowstyle }}
 \newcommand {\rowstyle}[1]{\gdef\currentrowstyle{#1} %
 #1\ignorespaces
 }
\newcommand{\tabhead}{\rowstyle{\bfseries}}

\begin{document}

\title{Investigating the Impact of Randomness on Reproducibility in Computer Vision: A Study on Applications in Civil Engineering and Medicine\\
\IEEEcompsocitemizethanks{This study was funded by a PhD grant from the DFG Research Training Group 2535 Knowledge-and data-based personalization of medicine at the point of care (WisPerMed).}}

\author{\IEEEauthorblockN{Bahadır Eryılmaz}
\IEEEauthorblockA{
\textit{University Hospital Essen}\\
Essen, Germany \\
0009-0002-8743-4751}
\and
\IEEEauthorblockN{Osman Alperen Koraş}
\IEEEauthorblockA{
\textit{University Hospital Essen}\\
Essen, Germany \\
0009-0006-6490-3139}
\and
\IEEEauthorblockN{Jörg Schlötterer}
\IEEEauthorblockA{
\textit{University of Marburg \& of Mannheim}\\
Marburg \& Mannheim, Germany \\
0000-0002-3678-0390}
\and
\IEEEauthorblockN{Christin Seifert}
\IEEEauthorblockA{
\textit{University of Marburg}\\
Marburg, Germany \\
0000-0002-6776-3868}
}

\maketitle

\begin{abstract}
    Reproducibility is essential for  scientific research. However, in computer vision, achieving consistent results is challenging due to various factors.
    One influential, yet often unrecognized, factor is CUDA-induced randomness.
    Despite CUDA's advantages for accelerating algorithm execution on GPUs, if not controlled, its behavior across multiple executions remains non-deterministic. While reproducibility issues in ML being researched, the implications of CUDA-induced randomness in application are yet to be understood. Our investigation focuses on this randomness across one standard benchmark dataset and two real-world datasets in an isolated environment.
    Our results show that CUDA-induced randomness can account for differences up to 4.77\% in performance scores.
    We find that managing this variability for reproducibility may entail increased runtime or reduce performance, but that disadvantages are not as significant as reported in previous studies.
\end{abstract}
	
\begin{IEEEkeywords}
deep learning, reproducibility, randomness, computer vision, determinism
\end{IEEEkeywords}

\label{src:introduction}
\section{Introduction}

The reproducibility crisis in machine learning is a growing concern that questions the reliability and validity of reported research findings~\cite{KAPOOR2023100804}. One survey shows that not all researchers are aware of this problem~\cite{pham2020problems}. This issue stems from the difficulty in replicating results due to various unknown and poorly understood factors, including but not limited to differences in data preparation, algorithmic details and computational environments. Deep learning architectures, as they are widely used in computer vision, with their complex, multi-layered neural networks~\cite{PIRAMUTHU1994509}, further impede reproducibility, as these models often involve numerous hyperparameters and training details that can lead to significant variability in results.

However, even in tightly controlled training environments, different training runs can lead to models with different weights and performances due to CUDA-induced randomness~\cite{zhuang2022randomness} caused by non-deterministic implementations of certain operations, differences in floating-point arithmetic precision, and the parallel execution order of operations, which may not be consistent across runs or different hardware setups. 

While it is possible to compare the performance of different modes on an appropriate evaluation dataset, the significance of such comparisons is limited to the specific checkpoints evaluated. The presence of CUDA-induced randomness, alongside other sources, complicates the attribution of performance differences between models to their architectural or algorithmic differences, since random factors alone can account for significant variances in scores~\cite{snapp2021synthesizing}. To conclusively assess whether one machine learning algorithm outperforms another, it would be necessary to train multiple models to sample the space induced by that algorithm~\cite{pham2020problems}. However, this approach is often impractical for a single study, especially given the trend towards larger models that demand increasing substantial computational resources.

Despite these concerns, the runtime improvements gained through the parallelization capabilities of CUDA GPUs indicates that their use will remain indispensable in the foreseeable future~\cite{JEON2021167, hooker2021hardware}. This reality emphasizes the importance of understanding CUDA-induced randomness and its impact on model performance in real-world applications. After all, a lack of such insight could lead researchers to overestimate or underestimate the capabilities of machine learning algorithms, which in turn could misdirect the efforts of the research community.

We investigate the effects of CUDA-induced randomness, its implications on reproducibility, and its broader implications on real-world computer vision applications.

\label{src:02_related_work}
\section{Related Work}

Goodman et al.~\cite{goodman:2016ca} provide a foundational definition of reproducibility, framing the standards we adopted in our study. 
Raste et al.~\cite{Raste}  analyzed the effects of randomness in model training and dataset partitioning, but did not investigate CUDA randomness in their setup.
Chen et al.~\cite{chenetal} detail the challenges inherent in reproducing deep learning results and showcase their solutions through empirical case studies. The authors emphasize CUDA randomness and employed various strategies to assess the impact of GPU execution-related randomness. 
Our objectives align with those of~\cite{chenetal} and our study employs more recent techniques to achieve fully deterministic results. Furthermore, we investigate the sensitivity of the output with respect to different optimizers and random seeds across multiple domains.
Scardapane et al.~\cite{scardapane2017randomness} provide  a comprehensive overview on randomness in deep learning, detailing the complexities and applications of randomness.  In a broader context, Dirnagl~\cite{dirnagl2019rethinking} considered the problem of reproducing any scientific work and investigated this across multiple domains, emphasizing the multifaceted nature of this problem.
Pham et al.~\cite{pham2020problems} studied the variance of performance of deep learning systems. Their results show substantial performance variance and a considerable knowledge gap among researchers about these inconsistencies.
Chou et al.~\cite{chou2020deterministic} promoted deterministic execution on GPU platforms, citing its benefits for reproducibility. Later, a benchmark study by Zhuang et al.~\cite{zhuang2022randomness} presented tooling insights to manage randomness, focusing on the interplay between algorithmic-level factors and implementation-level factors. They state that deterministic training can introduce significant overhead.

\label{src:03_methodology}
\section{Methodology}

For our empirical reproducibility investigation, we consider various hyperparameter configurations across different domains, conducting one fully deterministic and multiple randomized training runs. For the latter, we tightly control all sources of  randomness besides CUDA-induced randomness caused by non-deterministic implementations of certain operations. 

\begin{figure}[t]
    \centering
    \includegraphics[width=\columnwidth]{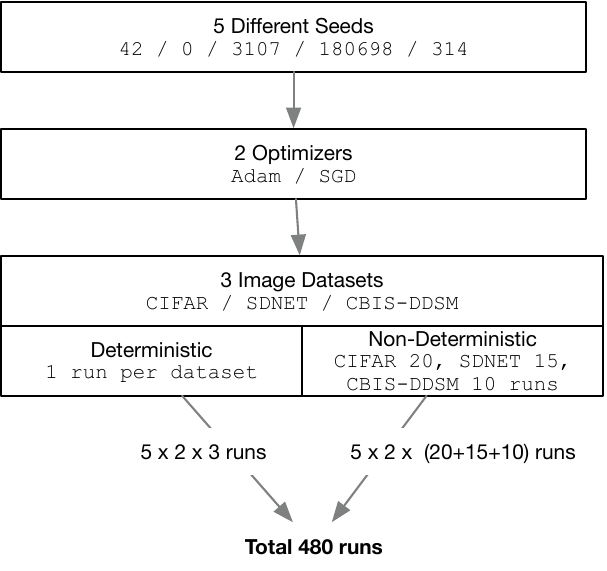}
    \caption{Experimentation setup. 20, 15 and 10 runs for CIFAR, SDNET and CBIS-DDSM, respectively per seed and optimizer in the non-deterministic setup, 480 runs in total.}
    \label{fig:runs_exp}
\end{figure}

To mitigate randomness in deep learning applications, we fix seeds to five different random values. This practice, while limiting randomness up to a certain point, ensures more consistent results for random operations such as weight initialization, data shuffling, and data augmentation~\cite{Ji_2023_WACV}. To investigate the role of optimizers, we use the two widely used optimizers ADAM~\cite{adamopt} and SGD with momentum~\cite{sgdmomopt}. We then study the sensitivity of these different configurations to CUDA-induced randomness. An overview of the setup is shown in Fig.~\ref{fig:runs_exp}.
      
As depicted in Fig.~\ref{fig:runs_exp}, CUDA provides two settings: nondeterministic and deterministic. We conducted experiments on three datasets: CIFAR-10~\cite{krizhevsky2009learning}, SDNET2018~\cite{maguire2018sdnet2018}, and CBIS-DDSM~\cite{spanhol2015dataset}, with 20, 15, and 10 runs respectively. Each run was performed using fixed seed configurations under nondeterministic CUDA settings. Additionally, each fixed seed configuration was also tested once under fully deterministic settings. Consistency in the experimental framework was ensured by maintaining the same libraries, their versions, hardware, and other environmental factors that could affect randomness. Furthermore, the influence of two commonly used optimizers on the outcomes was evaluated. Overall, we evaluated a total of 480 experimental runs across the three datasets.

CIFAR-10~\cite{krizhevsky2009learning} is a standard computer vision dataset with 60,000 color images across 10 classes. SDNET2018~\cite{maguire2018sdnet2018} is a real-world dataset with 56,000 labeled images for concrete crack detection. CBIS-DDSM~\cite{cbisddsm} is another real-world dataset from a different domain, featuring around 10,000 mammography images categorized into benign and malignant cases. 
We selected those datasets based on availability, citation frequency, and relevance to our research.

For model architectures, we employ ResNet~\cite{he2015deep}, PreActResNet~\cite{He2016}, and MobileNet~\cite{DBLP:journals/corr/HowardZCKWWAA17} for image classification tasks due to their effectiveness and community recognition. ResNet tackles training challenges in deep networks with ``residual blocks'', while PreActResNet optimizes performance through pre-activation integration. MobileNet, designed for resource-constrained devices, ensures efficiency without performance compromise. These choices were made considering efficacy, ease of implementation, and alignment with our research objectives.

All experiments were executed on a High-Performance Computer (HPC) infrastructure at the Institute for AI in Medicine~\cite{Schmidt2021_ieeecogmi_essen-medical-computing-platform} using SLURM~\cite{yoo2003slurm}, with monitoring facilitated by Weights \& Biases (W\&B)~\cite{wandb}.

\label{src:04_experiment_results}
\section{Results}

Table \ref{tab:optimizer_performance} summarizes the worst and best performance metrics  out of a total of 480 runs obtained across the three tasks at hand. The numbers show the impact of both the random seed and CUDA related randomness. 
\begin{table}[t]
    \caption{Aggregated optimizer performance metrics for every run each  task with maximum and minimum values with respective evaluation method.}
    \label{tab:optimizer_performance}
        \centering
    \small 
        \begin{tabular}{@{}lcccccc@{}}
        \toprule
        \textbf{Dataset} & \textbf{Metric} & \multicolumn{2}{c}{\textbf{ADAM (\%)}} & \multicolumn{2}{c}{\textbf{SGD (\%)}} \\
        \cmidrule(lr){3-4} \cmidrule(lr){5-6}
        & & \textbf{Max} & \textbf{Min} & \textbf{Max} & \textbf{Min} \\
        \midrule
        CIFAR-10     & Acc. & 93.57 & 91.11 & 95.12 & 94.41 \\
        SDNET2018 & F1   & 94.20 & 93.33 & 93.52 & 91.95 \\
        CBIS-DDSM & AUC  & 80.49 & 76.88 & 79.35 & 73.73 \\
        \bottomrule
    \end{tabular}

\end{table}
We evaluated the performance of different configurations on three distinct applications: classical CIFAR-10 benchmark, concrete crack detection with the SDNET2018 dataset, and medical imaging with the CBIS-DDSM dataset. 

Our results for CIFAR-10 align with \cite{gupta2021adam},with Stochastic Gradient Descent (SGD)\cite{sgdmomopt} converging to more similar accuracies than Adaptive Moment Estimation (ADAM)\cite{adamopt}. 
Our results are also consistent with \cite{He2016}, even though our approach utilized a simpler network architecture. 

For concrete crack detection, we report F1-score, as performance differences between optimizers were more pronounced. 
In the concrete crack detection case, our methodology shows improvement over the foundational work by \cite{dorafshan2018sdnet2018}, indicating the effectiveness of the chosen model in real-world scenarios.

For the medical imaging analysis using the CBIS-DDSM dataset, we report AUC \cite{bradley1997use} scores, a critical metric in this domain. Our findings align with those presented in the work of \cite{DBLP:journals/corr/abs-2002-07613}, albeit without employing GMIC (Globally-aware Multiple Instance Classifier) \cite{shen2019globally}, which is known for its low-memory consumption while enabling higher resolution. 

Table \ref{tab:combined_results} shows detailed results from all configurations, highlighting the variability in performance that can occur if CUDA randomness is not controlled. 
In particular, in the CBSI-DDSM task, a clear gap of a maximum difference of 4.77\% between the deterministic and non-deterministic runs with a random seed of 3407 and the SGD optimizer is noticeable.

\begin{table}[ht]
\centering
  \caption{Comparison of non-deterministic means ($\overline{\text{ND}}$) and deterministic (D) runs for CIFAR-10, SDNET, and CBIS-DDSM datasets across various configurations. The results show the performance metrics (Accuracy for CIFAR-10, F1-Score for SDNET, and AUC for CBIS-DDSM) with their respective configurations and the percentage change ($\Delta\%$). The max difference (in \%) for each configuration is shown as well ($\Delta S$).}
\label{tab:combined_results}
\begin{tabular}{lclrrr}
  \toprule
  Dataset & Configuration & $\overline{\text{ND}}$ & D &  $\Delta\%$ & $\Delta S (\%)$ \\
  \midrule
  \multirow{10}{*}{\parbox{1.3cm}{CIFAR-10 \\ (Accuracy)}}
  & ADAM\textsubscript{0} & 92.57 & 92.19 & 0.419 & 0.92 \\
  & SGD\textsubscript{0} & 94.80 & 94.96 & -0.156 & 0.52 \\
  & ADAM\textsubscript{180698} & 93.07 & 92.22 & 0.924 & 0.89 \\
  & SGD\textsubscript{180698} & 94.78 & 94.93 & 0.158 & 0.49\\
  & ADAM\textsubscript{314} & 92.15 & 92.17 & -0.016 & 1.80 \\
  & SGD\textsubscript{314} & 94.80 & 94.71 & 0.097 & 0.5 \\
  & ADAM\textsubscript{3407} & 92.51 & 92.72 & -0.210 & 1.81 \\
  & SGD\textsubscript{3407} & 94.66 & 94.59 & 0.083 & 0.42 \\
  & ADAM\textsubscript{42} & 92.57 & 92.71 & -0.143 & 1.62\\
  & SGD\textsubscript{42} & 94.76 & 94.81 & -0.048 & 0.69\\
  \midrule
  \multirow{10}{*}{\parbox{1.3cm}{SDNET \\ (F1-Score)}}
  & ADAM\textsubscript{0} & 93.26 & 93.70 & -0.465 & 0.30\\
  & SGD\textsubscript{0} & 92.34 & 92.64 & -0.322 & 0.98 \\
  & ADAM\textsubscript{180698} & 93.45 & 93.82 & -0.386 & 0.41 \\
  & SGD\textsubscript{180698} & 92.24 & 92.12 & 0.138 & 0.37 \\
  & ADAM\textsubscript{314} & 93.17 & 93.14 & 0.033 & 0.45 \\
  & SGD\textsubscript{314} & 92.29 & 91.58 & 0.787 & 0.64\\
  & ADAM\textsubscript{3407} & 93.35 & 93.29 & 0.070 & 0.21 \\
  & SGD\textsubscript{3407} & 91.50 & 92.03 & -0.571 & 1.19\\
  & ADAM\textsubscript{42} & 93.30 & 93.37 & 0.066 & 0.66 \\
  & SGD\textsubscript{42} & 92.51 & 92.31 & 0.214 & 0.26 \\
  \midrule
  \multirow{10}{*}{\parbox{1.3cm}{CBIS-DDSM \\ (AUC)}}
  & ADAM\textsubscript{0} & 79.26 & 78.17 & 1.399 & 1.00 \\
  & SGD\textsubscript{0} & 76.69 & 77.35 & -0.854 & 3.86 \\
  & ADAM\textsubscript{180698} & 78.14 & 77.16 & 1.266 & 1.92\\
  & SGD\textsubscript{180698} & 76.42 & 77.73 & -1.681 & 2.81 \\
  & ADAM\textsubscript{314} & 77.75 & 79.36 & -2.022 & 3.22\\
  & SGD\textsubscript{314} & 77.18 & 77.70 & -0.674 & 4.40\\
  & ADAM\textsubscript{3407} & 79.69 & 78.77 & 1.172 & 2.46 \\
  & SGD\textsubscript{3407} & 77.23 & 76.56 & 0.876 & 4.77 \\
  & ADAM\textsubscript{42} & 78.38 & 79.65 & -1.596 & 2.13\\
  & SGD\textsubscript{42} & 77.24 & 76.95 & 0.364 & 2.23 \\
  \bottomrule
\end{tabular}
\end{table}
\begin{table*}[ht!]
\centering
\caption{Deterministic and non-deterministic performance comparison for the CBIS-DDSM dataset with two optimizers, ADAM and SGD, expressed in AUC score. The first column lists all the seed configurations. AUC$_{\text{det}}$ represents the deterministic run for each seed configuration. AUC$_{\text{n\_det}}$ represents the mean of non-deterministic runs for each seed configuration. $\Delta\%$ indicates the performance difference in percentage between the deterministic and non-deterministic runs. $\sigma_{\text{n-det(AUC)}}$ shows the standard deviation of all runs for the respective seed configuration with AUC scores, and $\sigma_{\text{n-det(F1)}}$ presents it for the F1-score.}
\label{tab:comparison_adam_sgd_f1}
\resizebox{\textwidth}{!}{%
\begin{tabular}{+c^c^c^c^c^c^c^c^c^c^c}
\toprule\tabhead
Seed & \multicolumn{5}{c}{ADAM} & \multicolumn{5}{c}{SGD} \\ 
\cmidrule(lr){2-6} \cmidrule(lr){7-11}
 & $\bar{AUC}_{\text{det}}$ & $\bar{AUC}_{\text{n-det}}$ & $\Delta\%$ & $\sigma_{\text{n-det(AUC)}}$ & $\sigma_{\text{n-det(F1)}}$ & $\bar{AUC}_{\text{det}}$ & $\bar{AUC}_{\text{n-det}}$ & $\Delta\%$ & $\sigma_{\text{n-det(AUC)}}$ & $\sigma_{\text{n-det(F1)}}$ \\
\otoprule
0 & 0.7817 & 0.7903 & 1.11 & 0.0037 & 0.0498 & 0.7735 & 0.7664 & -0.93 & 0.0141 & 0.0228 \\
180698 & 0.7716 & 0.7796 & 1.03 & 0.0057 & 0.0222 & 0.7773 & 0.7649 & -1.60 & 0.0090 & 0.0248 \\
314 & 0.7936 & 0.7835 & -1.27 & 0.0093 & 0.0302 & 0.7770 & 0.7737 & -0.44 & 0.0158 & 0.0197 \\
3407 & 0.7877 & 0.7940 & 0.80 & 0.0086 & 0.0244 & 0.7656 & 0.7698 & 0.54 & 0.0142 & 0.0190 \\
42 & 0.7965 & 0.7829 & -1.71 & 0.0066 & 0.0262 & 0.7695 & 0.7731 & 0.46 & 0.0082 & 0.0185 \\
\bottomrule
\end{tabular}%
}
\end{table*}

\begin{table}[ht!]
\centering
\caption{Deterministic and non-deterministic runtime comparison for the CBIS-DDSM dataset with two optimizers, ADAM and SGD, expressed in minutes. The first column lists all the seed configurations. $\mu_{\text{det}}$ represents the deterministic runtime for each seed configuration. $\mu_{\text{n-det}}$ represents the mean non-deterministic runtime for each seed configuration. $\Delta\%$ indicates the runtime difference in percentage between the deterministic and non-deterministic runs.}
\label{tab:runtime_comparison_adam_sgd}
\resizebox{\columnwidth}{!}{%
\begin{tabular}{+c^c^c^c^c}
\toprule\tabhead
Config & Optimizer & $\mu_{\text{det}}$ (min) & $\mu_{\text{n-det}}$ (min) & $\Delta\%$ \\
\midrule
\multirow{2}{*}{0} & ADAM & 343.63 & 319.85 & -6.92 \\
 & SGD & 398.45 & 358.88 & -9.93 \\
\midrule
\multirow{2}{*}{180698} & ADAM & 423.07 & 409.54 & -3.20 \\
 & SGD & 399.05 & 393.92 & -1.29 \\
\midrule
\multirow{2}{*}{314} & ADAM & 407.55 & 370.17 & -9.17 \\
 & SGD & 369.20 & 367.45 & -0.47 \\
\midrule
\multirow{2}{*}{3407} & ADAM & 414.48 & 382.11 & -7.81 \\
 & SGD & 364.17 & 303.99 & -16.53 \\
\midrule
\multirow{2}{*}{42} & ADAM & 360.46 & 355.44 & -1.39 \\
 & SGD & 366.48 & 376.31 & 2.68 \\
\bottomrule
\end{tabular}%
}
\end{table}

\subsection*{Performance Variance and Tradeoffs }

Our study employs distinct pipelines and metrics for each task, making direct performance comparisons across tasks challenging. However, by analyzing variances, we can assess task sensitivities to inherent randomness across the three distinct tasks. Additionally, for each seed and optimizer configuration, we have one fully deterministic configuration. By comparing the means of the non-deterministic runs with that particular fully deterministic run, we can get insights about the implications of CUDA randomness in terms of runtime and performance.

In analyzing the CBIS-DDSM task, as summarized in Table \ref{tab:comparison_adam_sgd_f1}, it is evident that the variability in performance metrics across different seeds and optimizers is noteworthy. 
For both, the SGD and the ADAM optimizer, we observe fluctuations in AUC scores, that are more pronounced in the ADAM optimizer. In particular, the ADAM optimizer at seed 0 shows the largest standard deviation in F1-score overall.
The impact of deterministic execution on performance varied, with decreases of up to 2\% for seed 314 and increases of up to 1.4\% for seed 0. Such findings highlight the stochastic nature of model training outcomes in relation to seed selection and optimizer choice. Additionally, we infer from Table~\ref{tab:runtime_comparison_adam_sgd} that runtime trade-offs from deterministic execution are generally small (once even in favor of the deterministic run) and at most incur a 16\% longer runtime.

\subsection*{T-Test with One-Sample Mean}
To evaluate if non-deterministic run performances significantly diverge from deterministic runs, we employed a t-test, leveraging the consistency of deterministic runs as the population mean. This approach is premised on the repeatability of deterministic runs, which are posited to accurately represent the "true" population mean. The hypotheses are succinctly framed as: \textbf{$H_0$ (Null)}: There is no significant difference between deterministic and non-deterministic runs, indicating the effect of CUDA randomness cannot be conclusively determined. \textbf{$H_1$ (Alternative)}: A significant difference suggests CUDA randomness does influence results, leading to the rejection of $H_0$. This methodology enables a direct assessment of CUDA randomness' impact on performance metrics.

\begin{table}[t]
  \centering
  \caption{Number of significant configurations (p $\leq$ 0.05, as determined by a t-test) from each dataset with respect to the optimizers, and average differences in performance between the significant configurations and all configurations.}
  \label{tab:summary_observations}
  \setlength{\tabcolsep}{4pt} 
  \renewcommand{\arraystretch}{1} 
  \small
  \begin{tabularx}{\columnwidth}{@{}lXXX@{}}
    \toprule
    & \textbf{Sig. Configs} & \textbf{Avg (Sig.)} & \textbf{Avg (All)} \\
    \midrule
    \textbf{CIFAR-10} \\
    SGD & 2/5 & 0.1465\% & 0.1042\% \\
    ADAM & 3/5 & 0.517\% & 0.3484\% \\
    \midrule
    \textbf{CBIS-DDSM} \\
    SGD & 1/5 & 1.68\% & 0.889\% \\
    ADAM & All & 1.491\% & 1.491\% \\
    \midrule
    \textbf{SDNET} \\
    SGD & 3/5 & 0.441\% & 0.406\% \\
    ADAM & 2/5 & 0.425\% & 0.204\% \\
    \bottomrule
  \end{tabularx}
\end{table}

The analysis of one-sample t-test results, as summarized in Table~\ref{tab:summary_observations}, indicates that CUDA randomness  affects the performance outcomes across the CIFAR-10, CBIS-DDSM, and SDNET datasets. For the CIFAR-10 dataset, statistically significant differences were observed in 50\% of the configurations, with ADAM showing a higher average deviation (0.517\% for significant runs, 0.3484\% overall) compared to SGD (0.1465\% for significant runs, 0.1042\% overall). In the CBIS-DDSM dataset, all ADAM configurations were significant, highlighting its sensitivity, whereas only one SGD configuration showed statistical significance, with average deviations of 1.68\% for significant runs and 0.889\% overall for SGD, and 1.491\% for both significant and overall runs for ADAM. The SDNET dataset showed a balance of significant outcomes among configurations for both optimizers, with ADAM's deviations at 0.425\% for significant runs and 0.204\% overall, versus SGD's 0.441\% for significant runs and 0.406\% overall. These findings underscore ADAM's higher sensitivity to CUDA randomness compared to SGD, reflected in the greater number of significant configurations and larger deviation percentages.

\subsection*{Similarities of Model Weights}

\begin{figure*}[tbp]
\begin{minipage}{0.02\textwidth}
\rotatebox{90}{CIFAR-10}
\end{minipage}%
\begin{minipage}{0.97\textwidth}
 \includegraphics[height=0.209\textwidth]{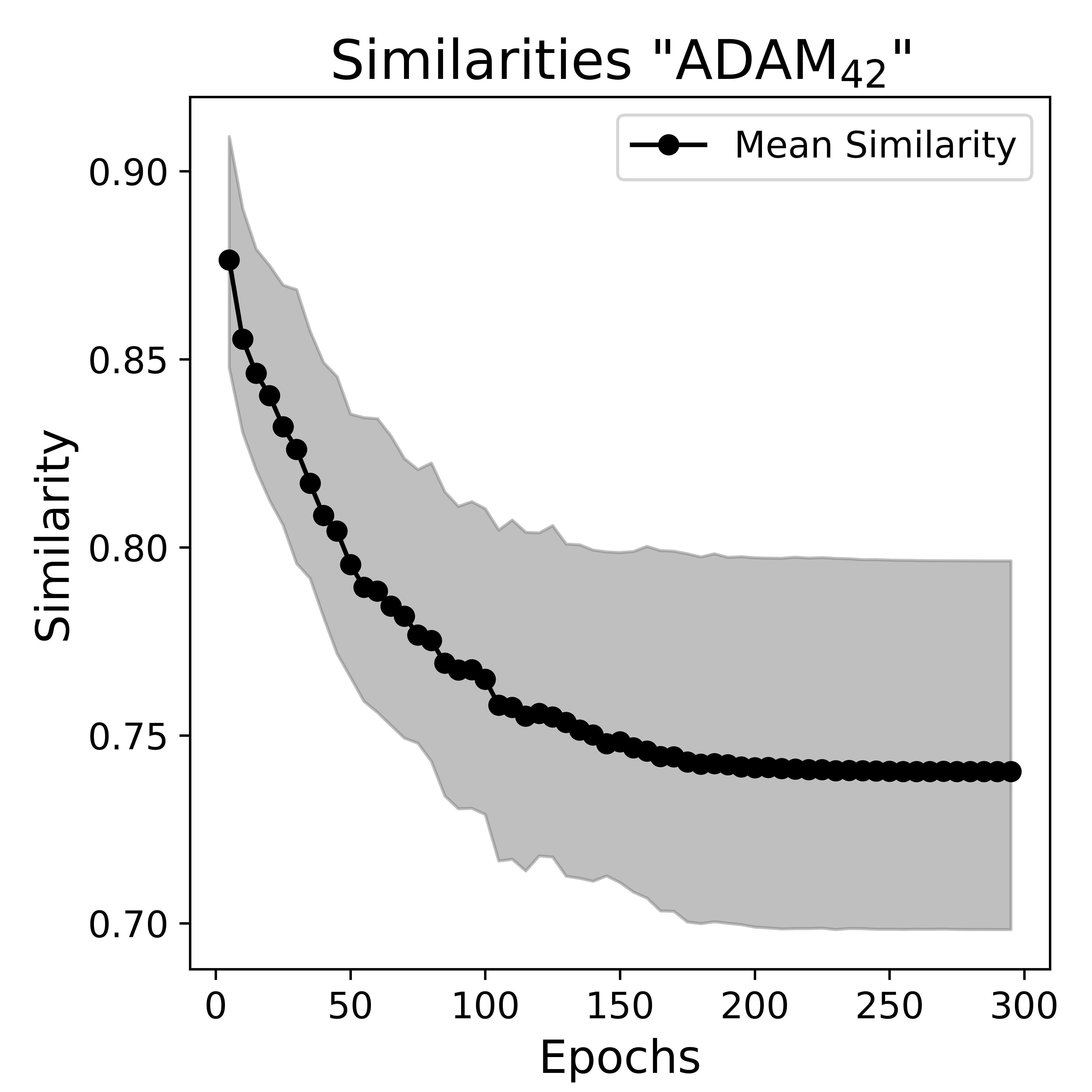}
\includegraphics[height=0.209\textwidth, trim={1cm 0cm 0cm 0cm}, clip]{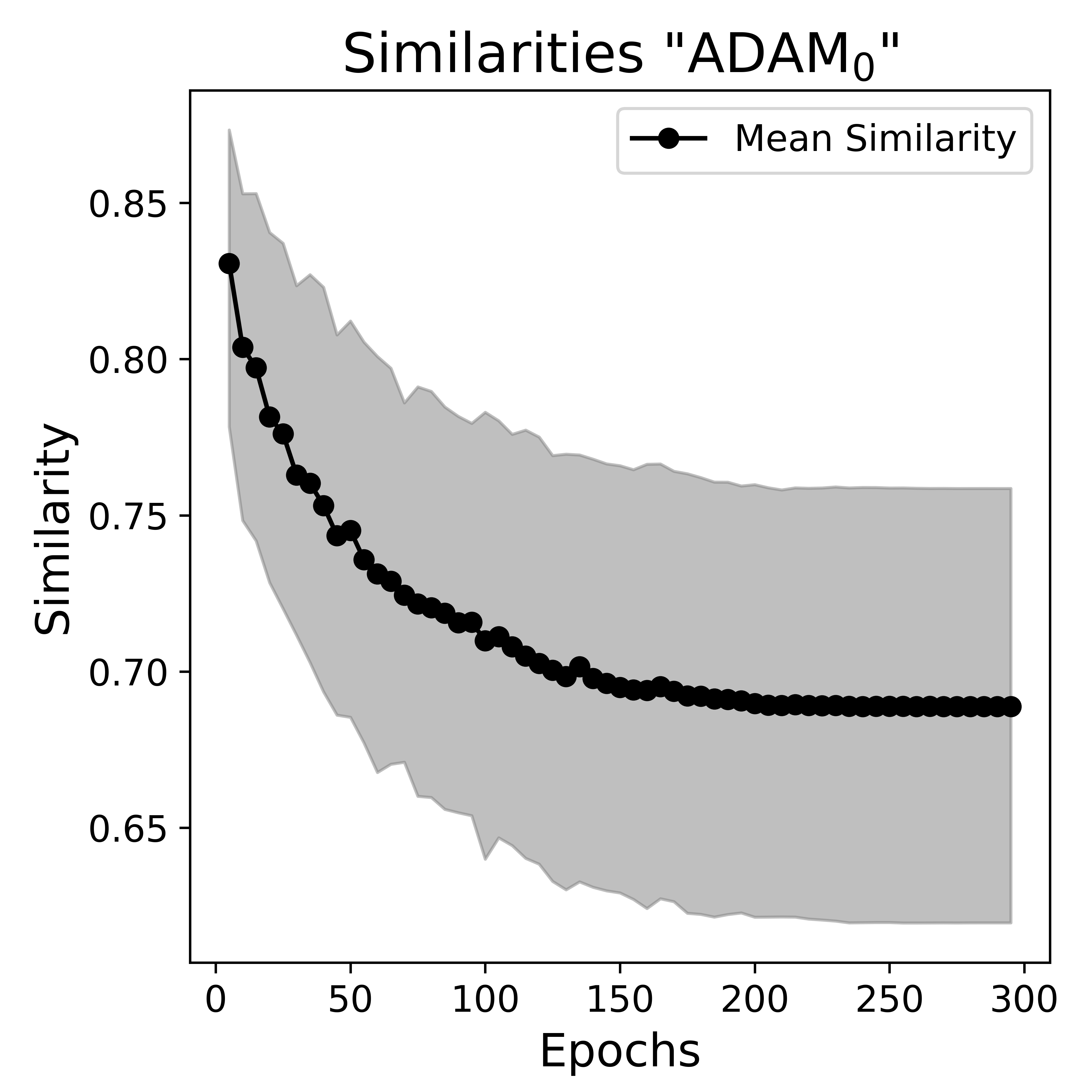}
\includegraphics[height=0.209\textwidth, trim={1cm 0cm 0cm 0cm}, clip]{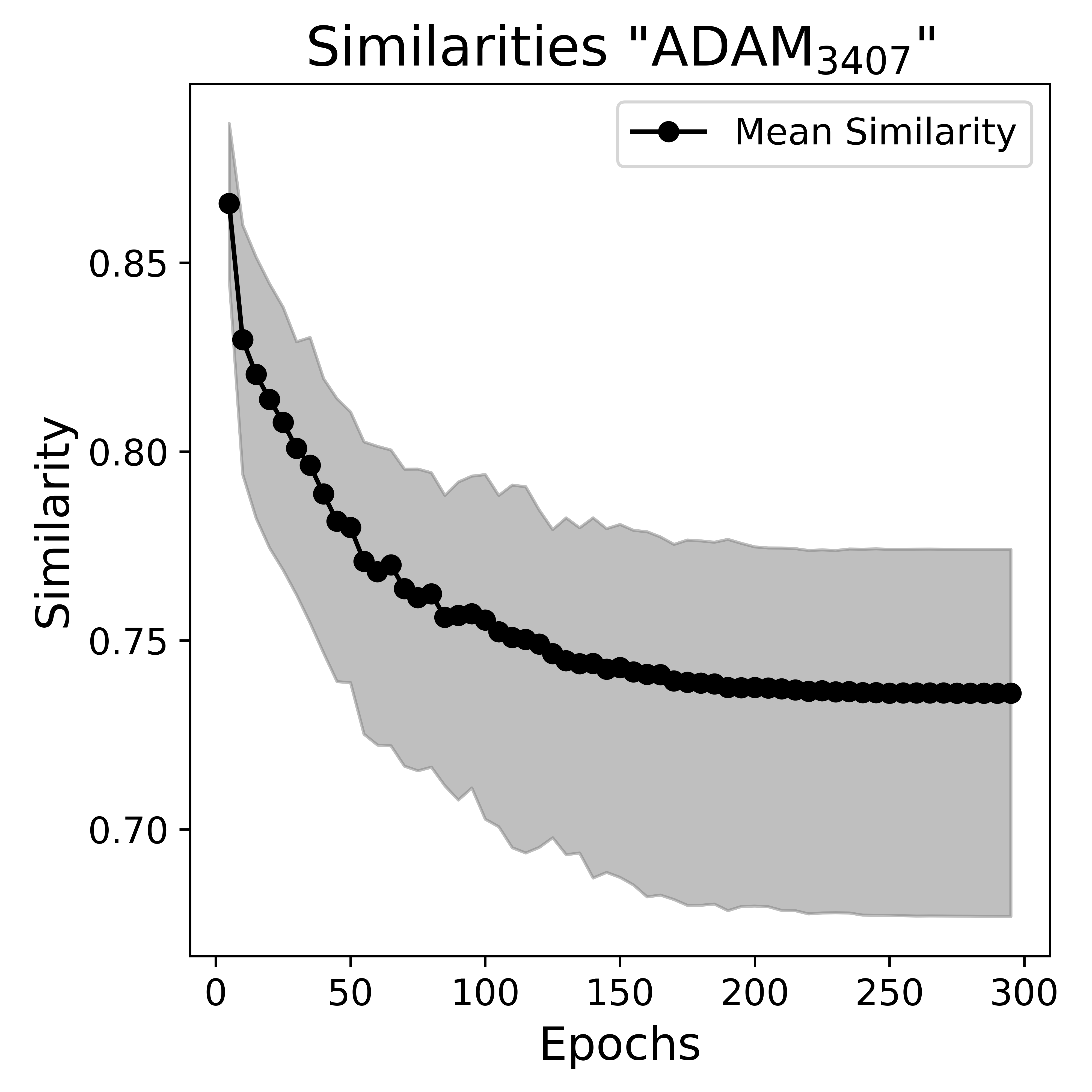}
\includegraphics[height=0.209\textwidth, trim={1cm 0cm 0cm 0cm}, clip]{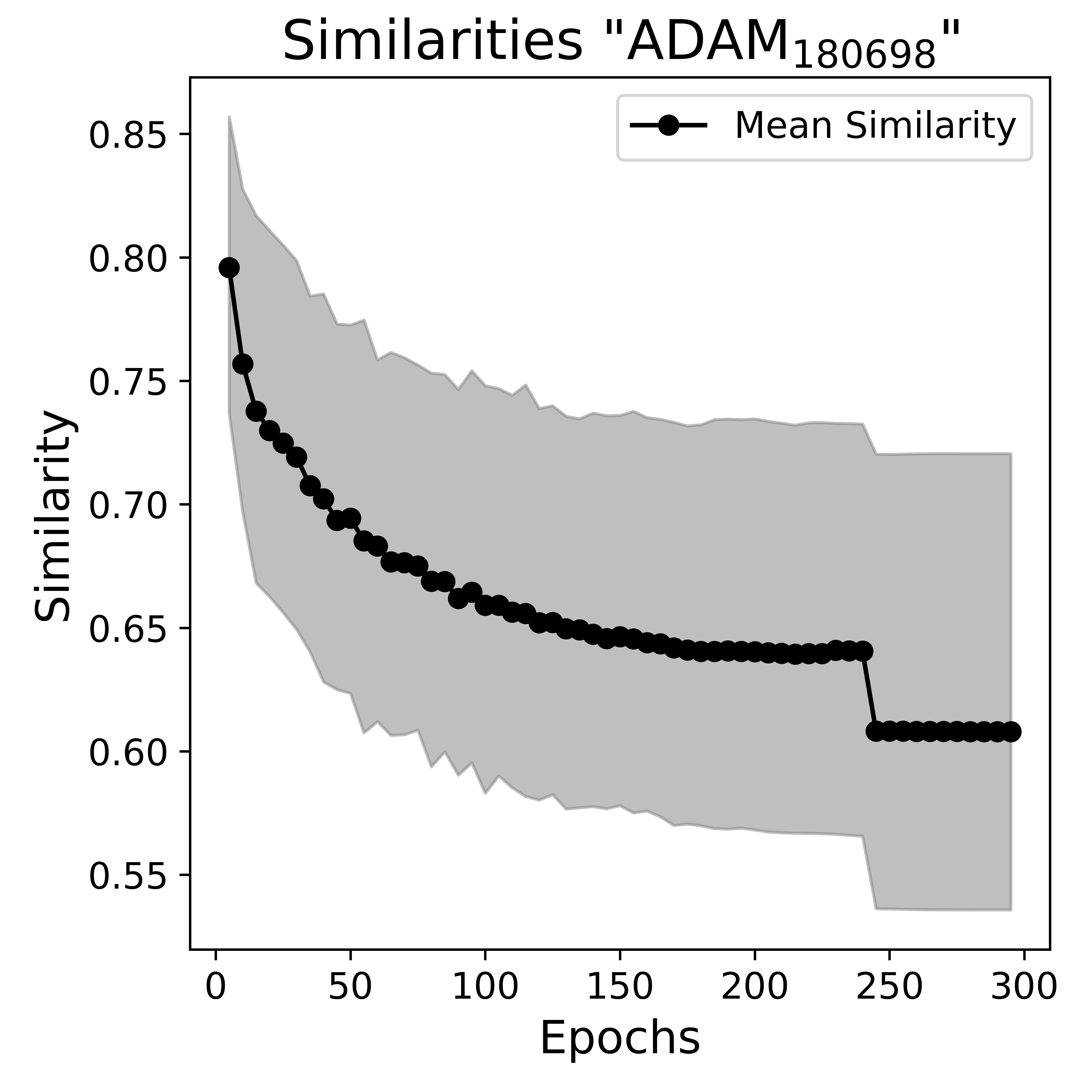}
\includegraphics[height=0.209\textwidth, trim={1.3cm 0cm 0cm 0cm}, clip]{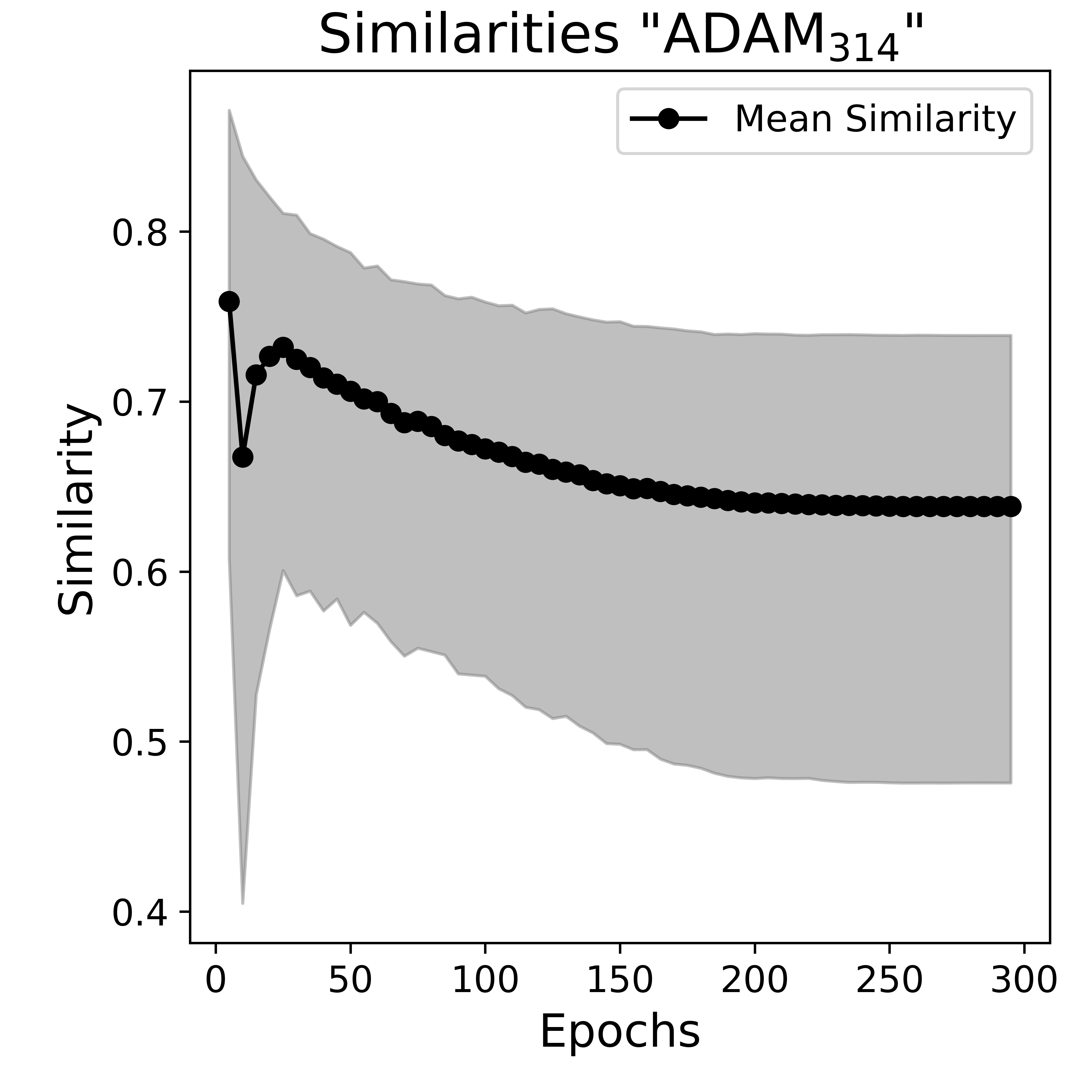}
\end{minipage}
\begin{minipage}{0.02\textwidth}
\rotatebox{90}{CBIS-DDSM}
\end{minipage}%
\begin{minipage}{0.97\textwidth}
 \includegraphics[height=0.216\textwidth]{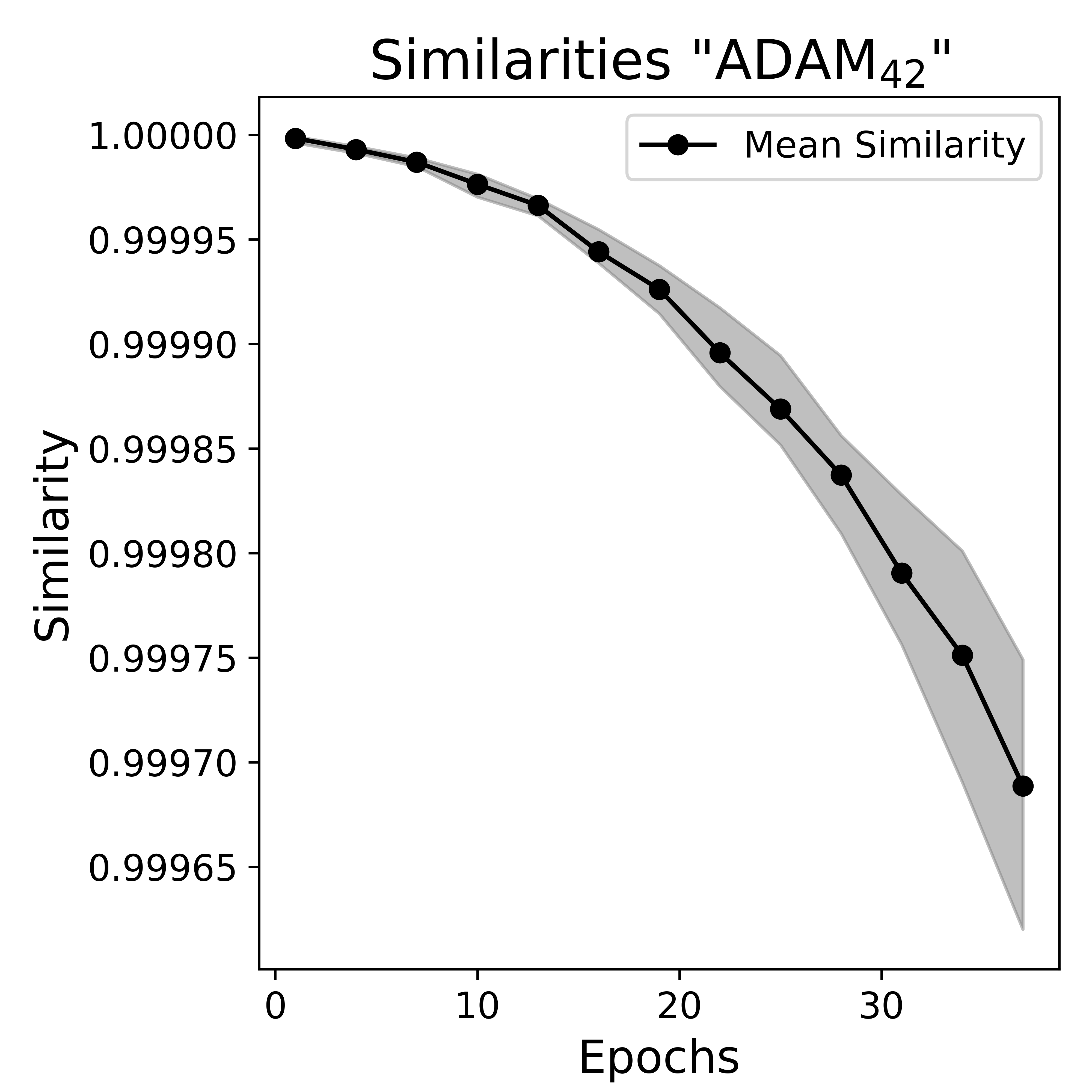}
\includegraphics[height=0.216\textwidth, trim={1.5cm 0cm 0cm 0cm}, clip]{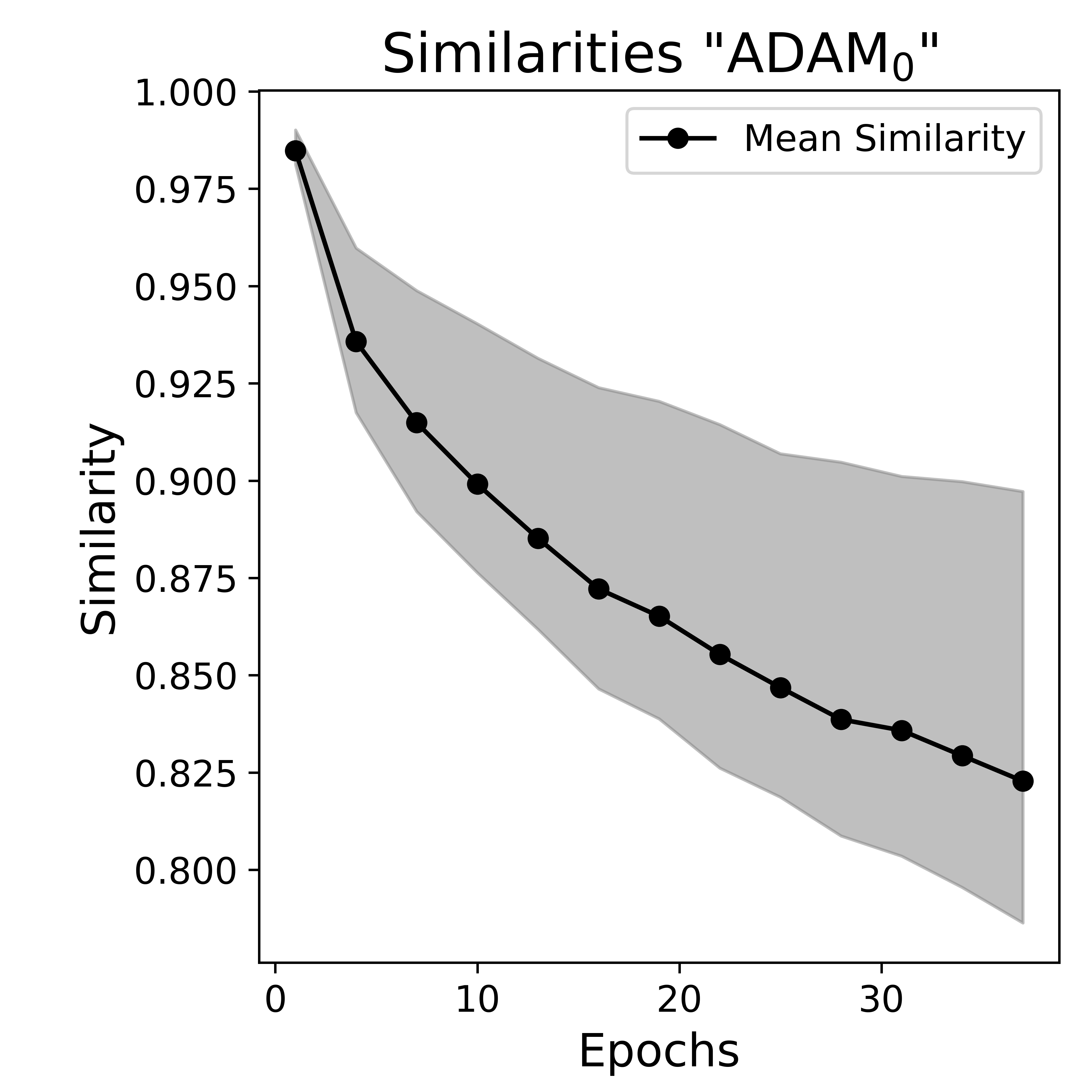}
\includegraphics[height=0.216\textwidth, trim={1.8cm 0cm 0cm 0cm}, clip]{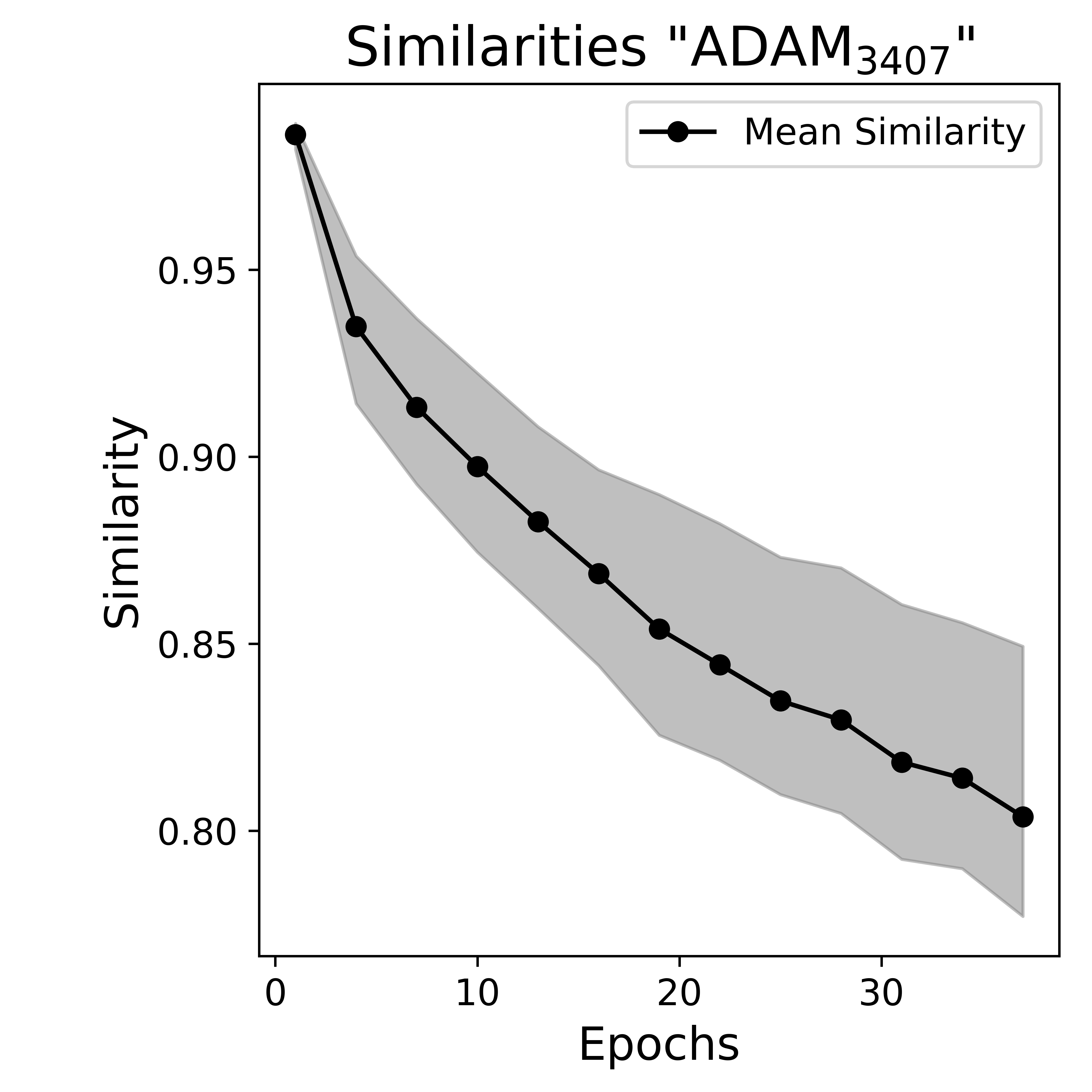}
\includegraphics[height=0.216\textwidth, trim={1.5cm 0cm 0cm 0cm}, clip]{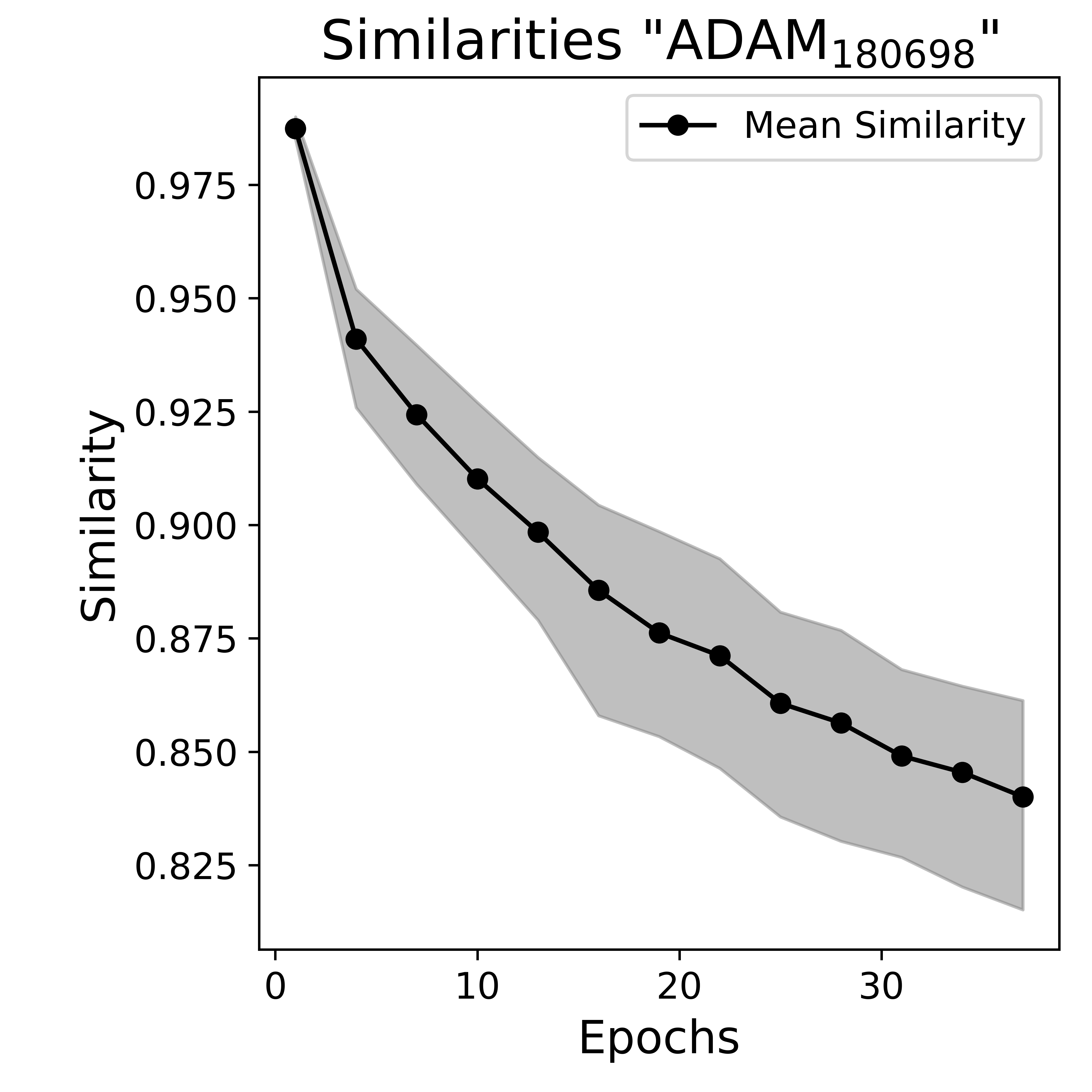}
\includegraphics[height=0.216\textwidth, trim={1.7cm 0cm 0cm 0cm}, clip]{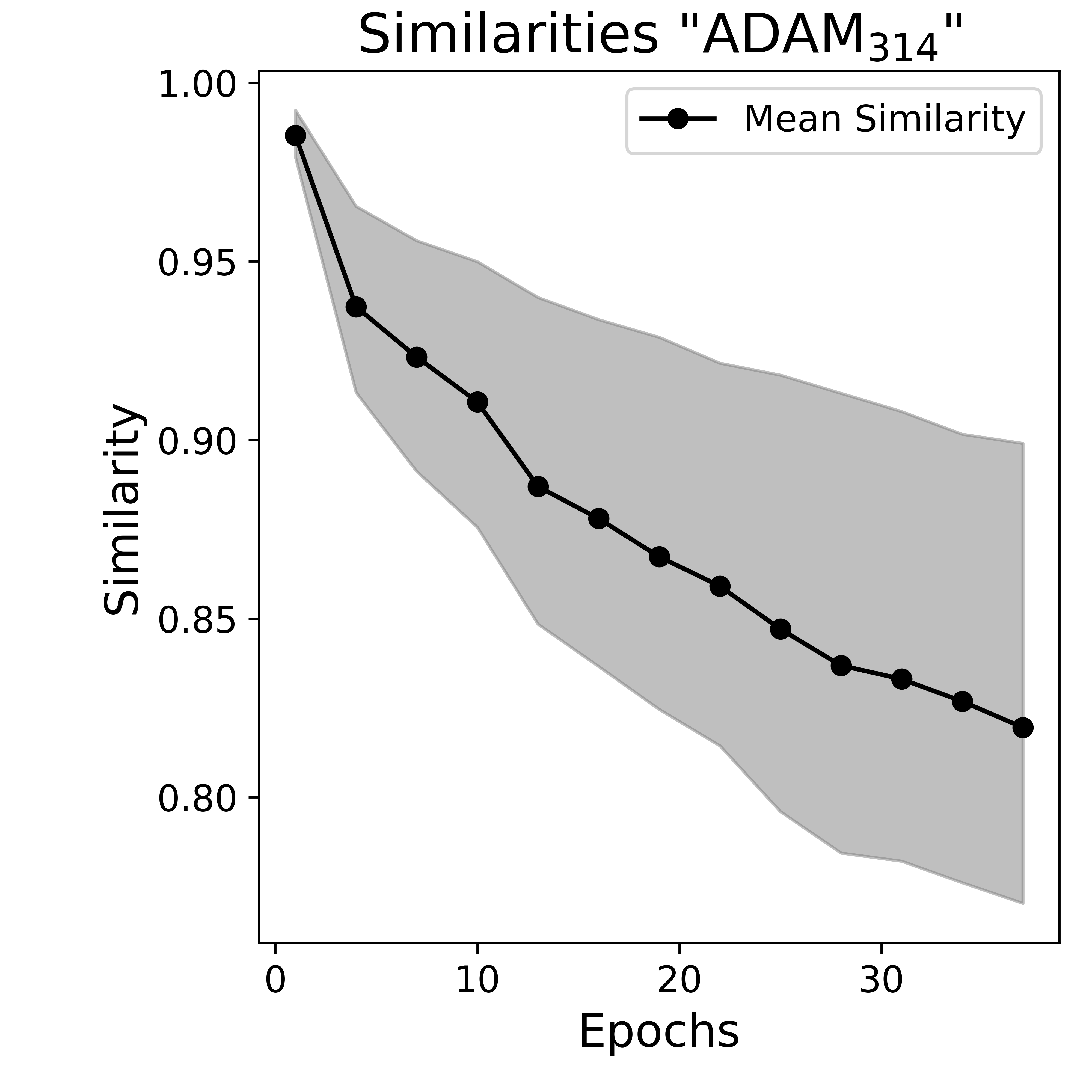}
\end{minipage}
\begin{minipage}{0.02\textwidth}
\rotatebox{90}{SDNET}
\end{minipage}%
\begin{minipage}{0.97\textwidth}
 \includegraphics[height=0.21\textwidth]{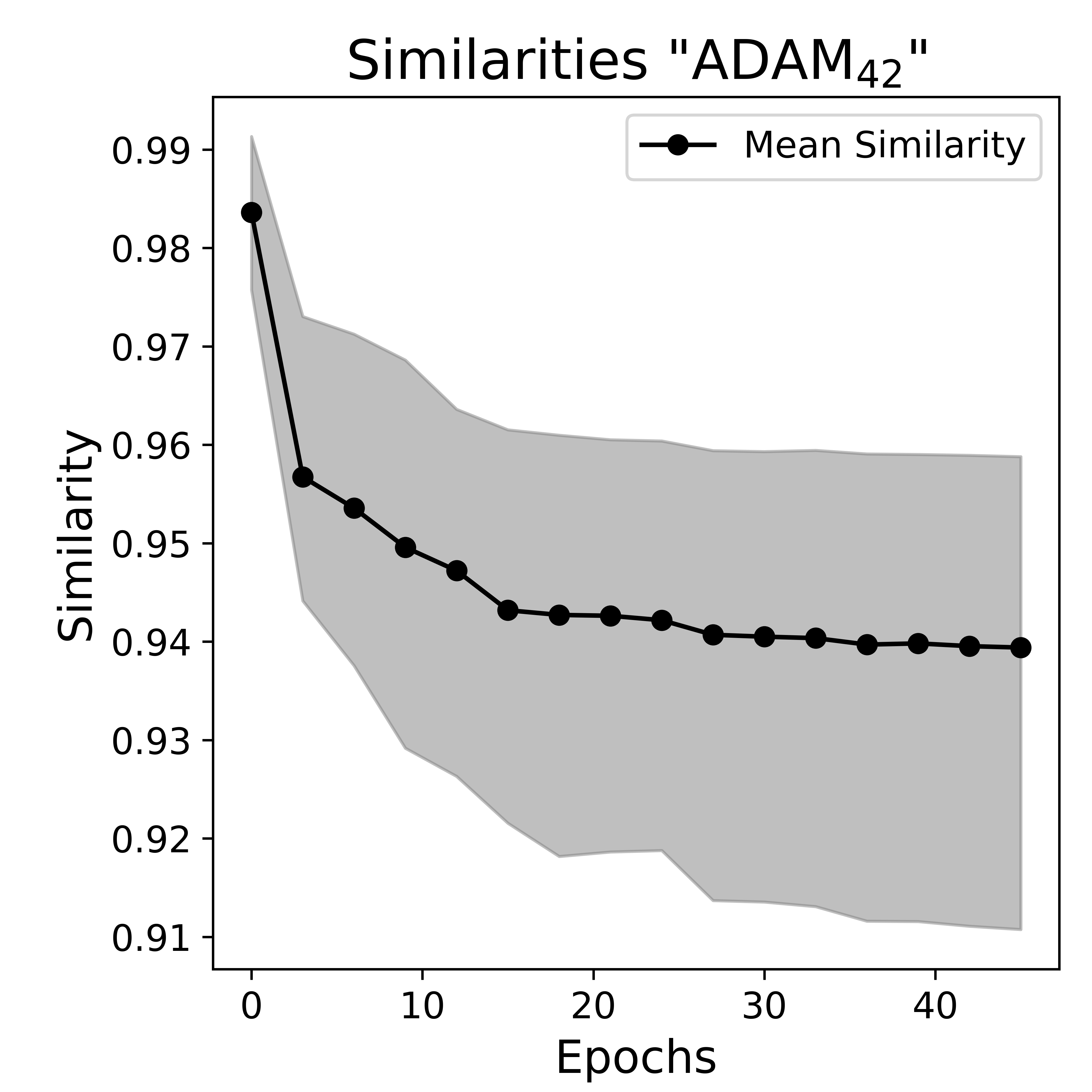}
\includegraphics[height=0.21\textwidth, trim={1cm 0cm 0cm 0cm}, clip]{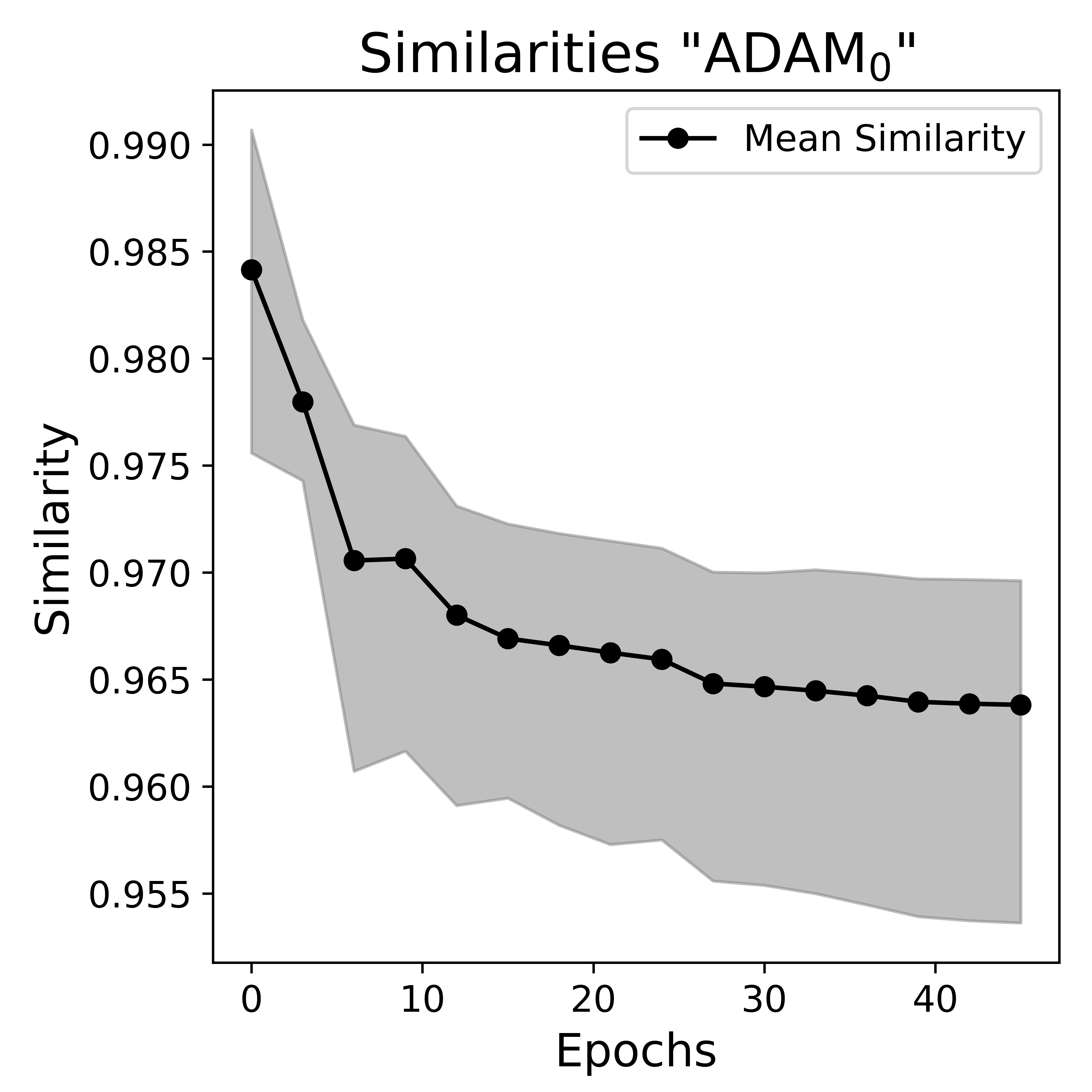}
\includegraphics[height=0.21\textwidth, trim={1.3cm 0cm 0cm 0cm}, clip]{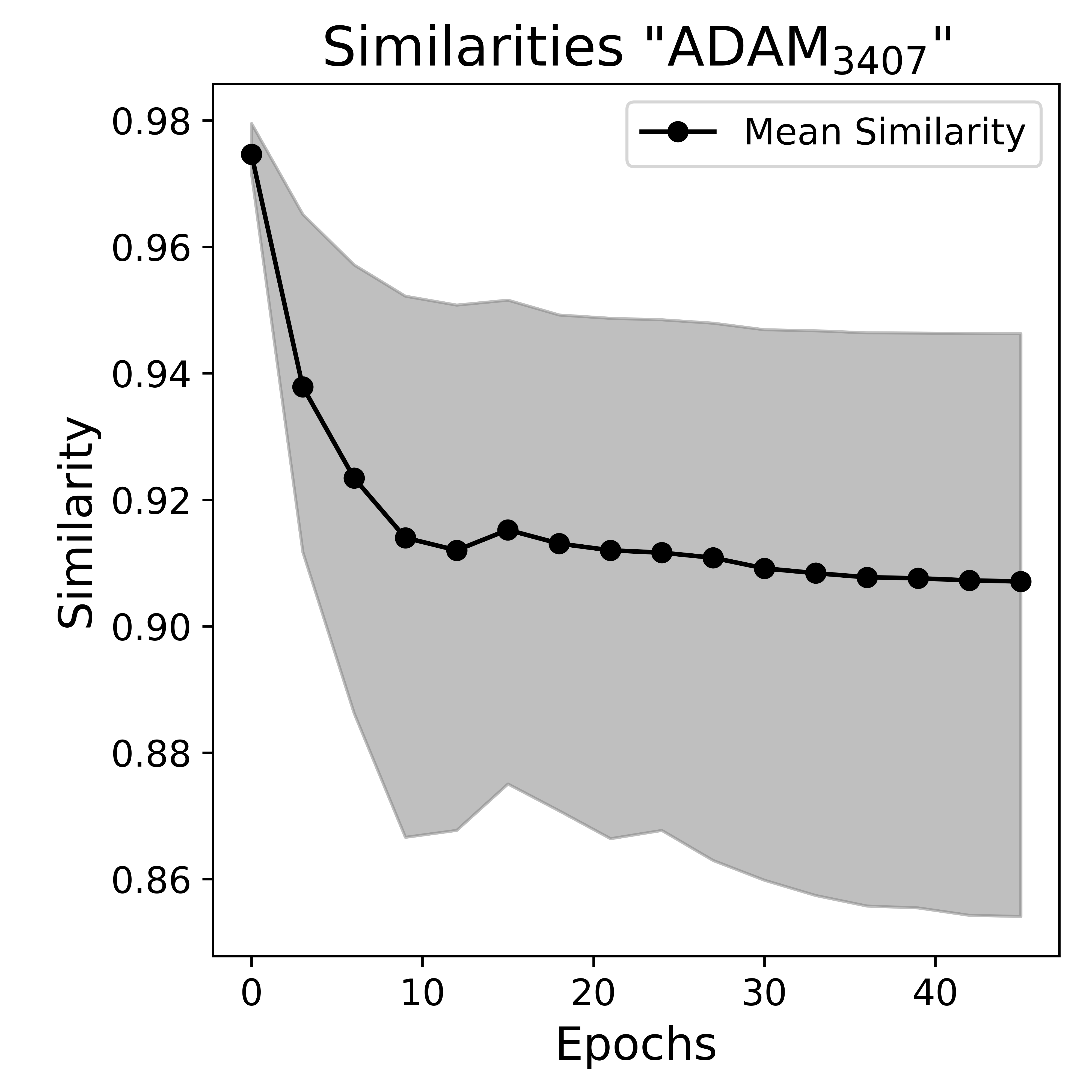}
\includegraphics[height=0.21\textwidth, trim={1cm 0cm 0cm 0cm}, clip]{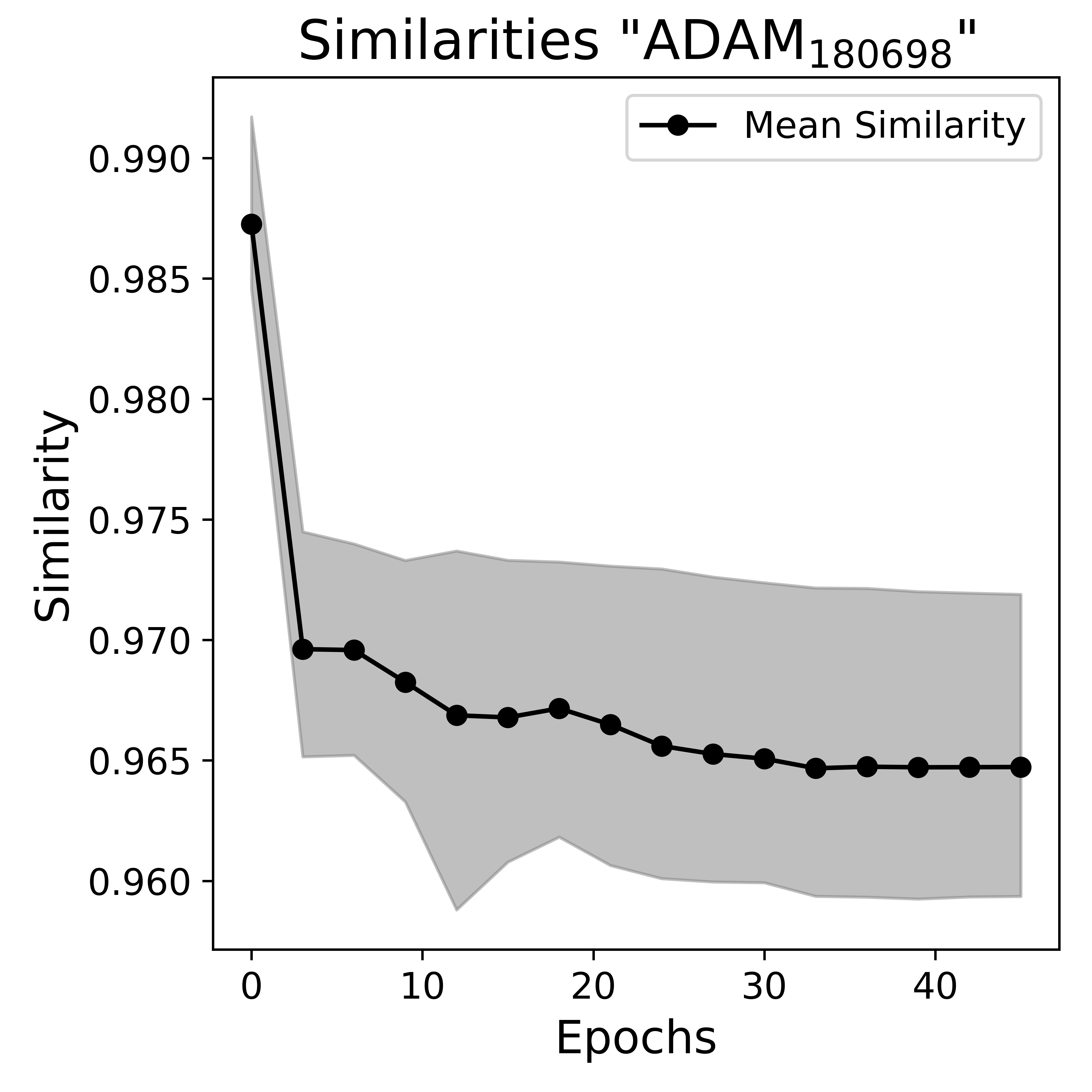}
\includegraphics[height=0.21\textwidth, trim={1.3cm 0cm 0cm 0cm}, clip]{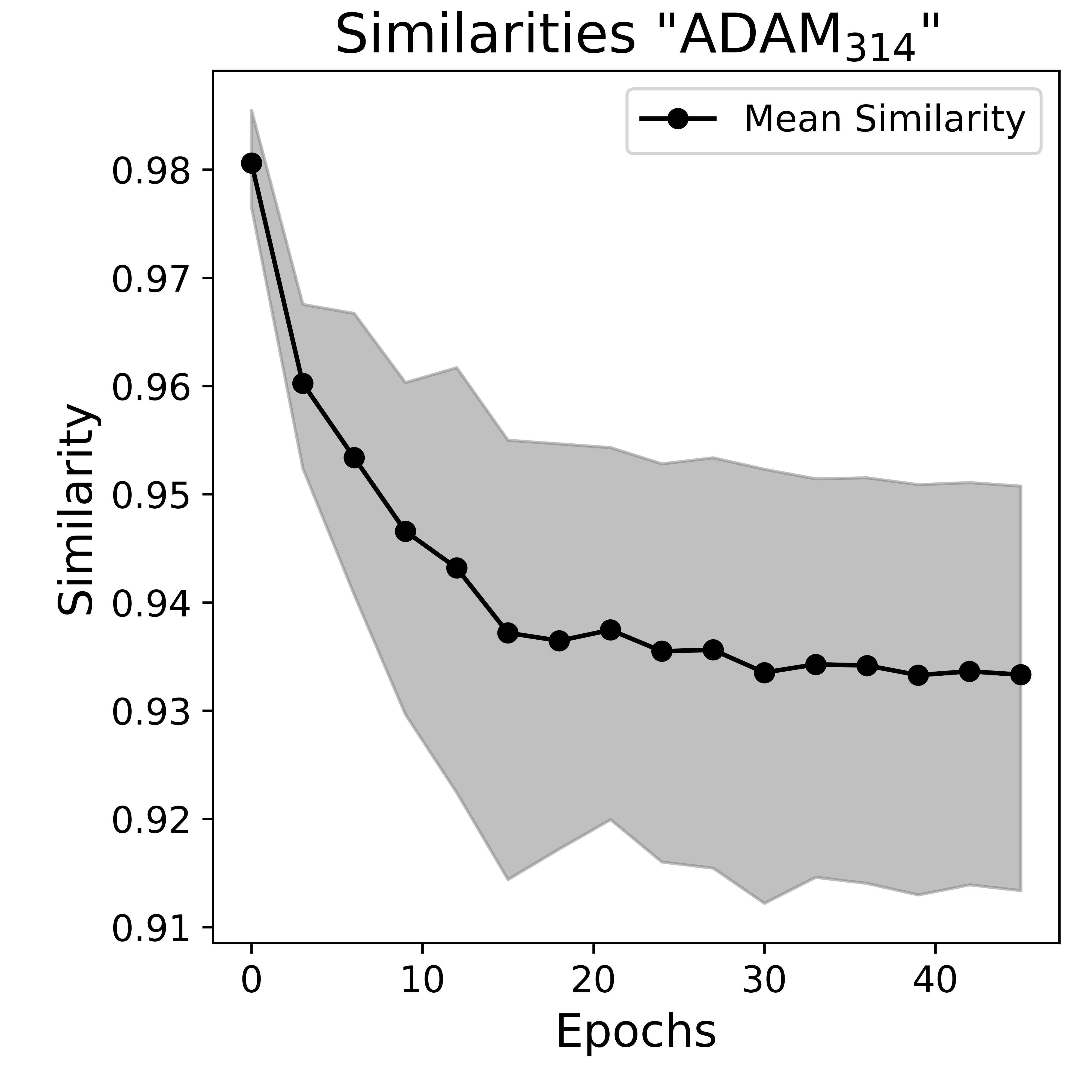}
\end{minipage}
  \caption{Cosine similarity of embeddings in the last linear layer of the models with respect to epoch number. Showing max, min, and mean cosine similarity values with Adam optimizer for each dataset (rows) and seed (column).}
  \label{fig:similarities_adam}
\end{figure*}
\begin{figure*}[tbp]
\begin{minipage}{0.02\textwidth}
\rotatebox{90}{CIFAR-10}
\end{minipage}%
\begin{minipage}{0.97\textwidth}
 \includegraphics[height=0.21\textwidth]{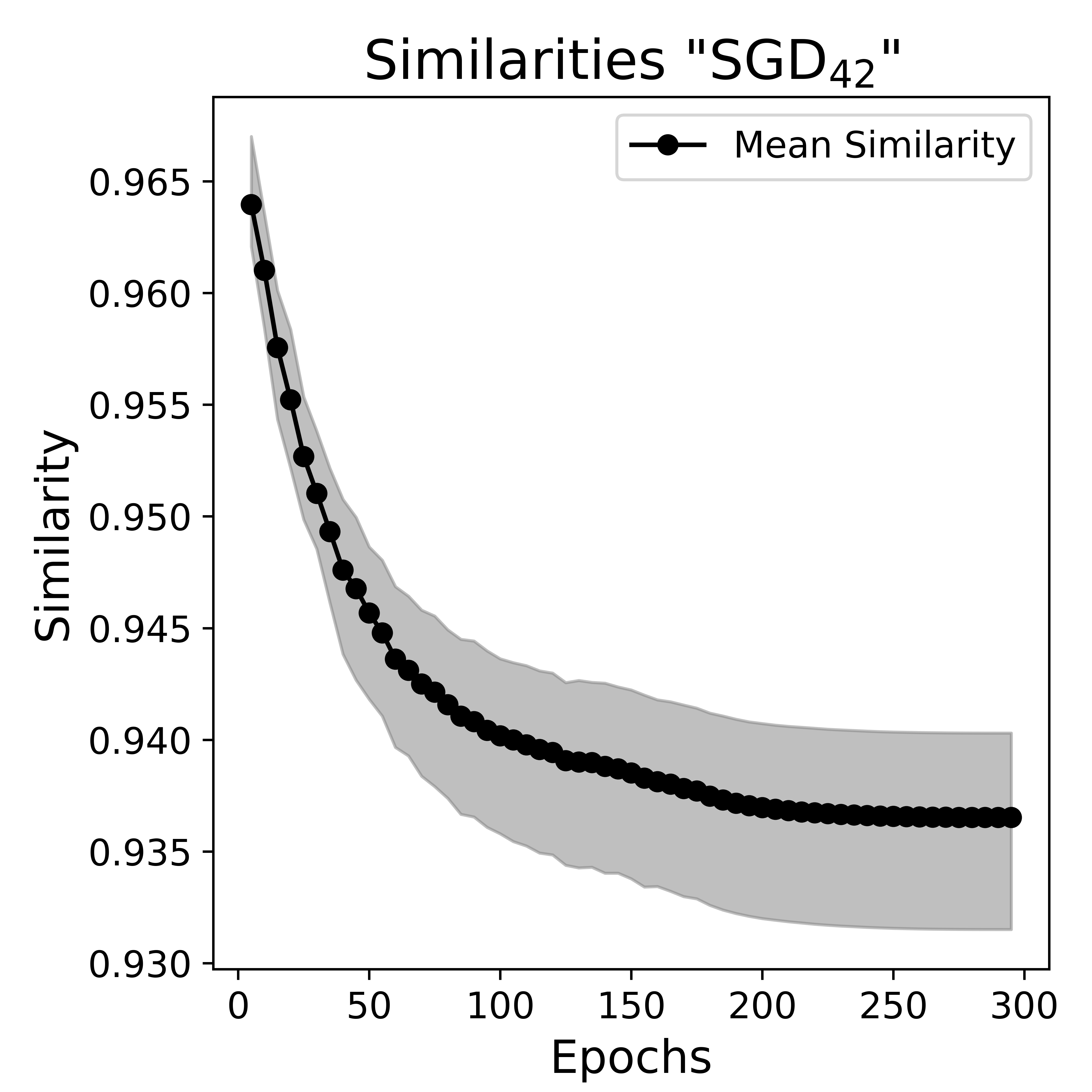}
\includegraphics[height=0.21\textwidth, trim={1.2cm 0cm 0cm 0cm}, clip]{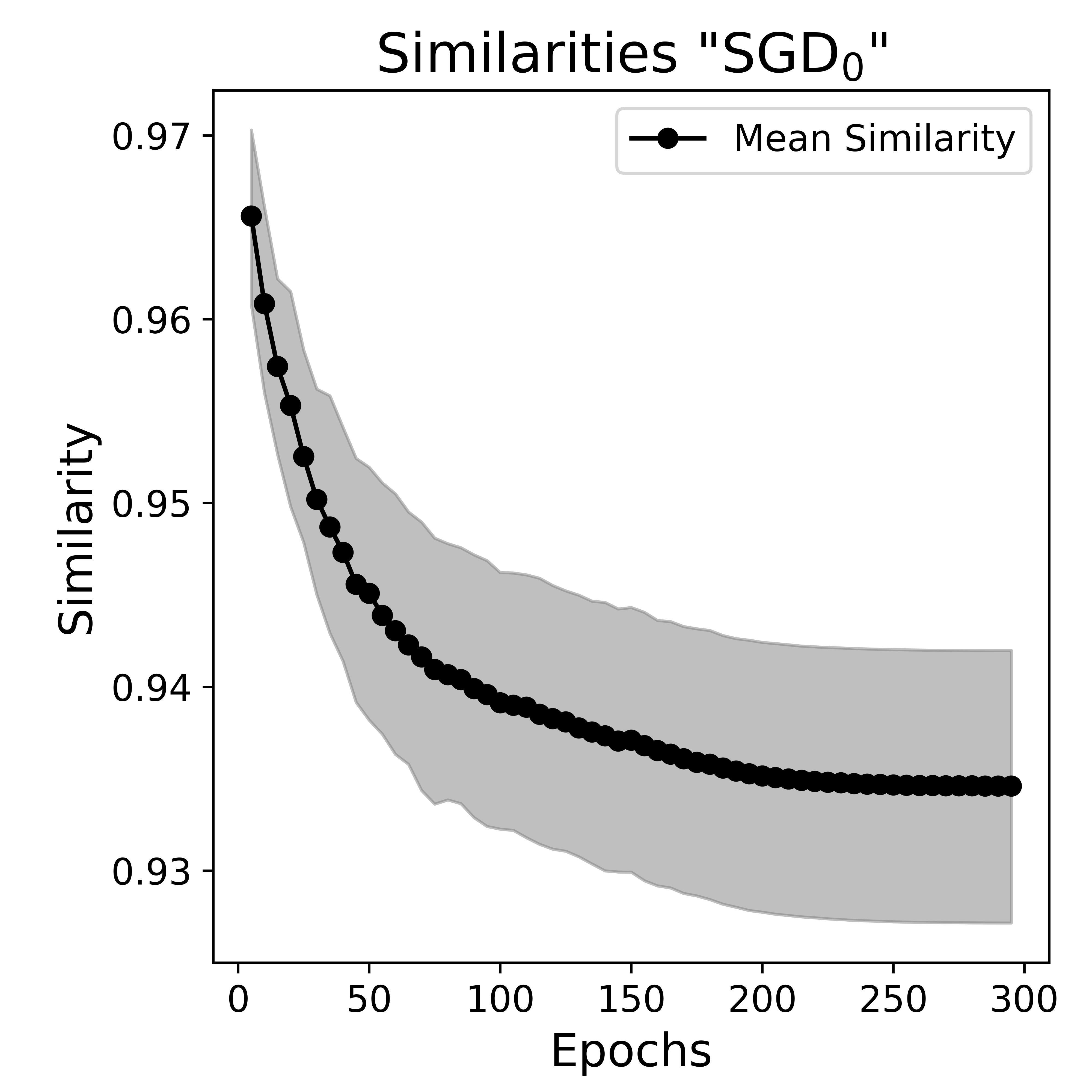}
\includegraphics[height=0.21\textwidth, trim={1cm 0cm 0cm 0cm}, clip]{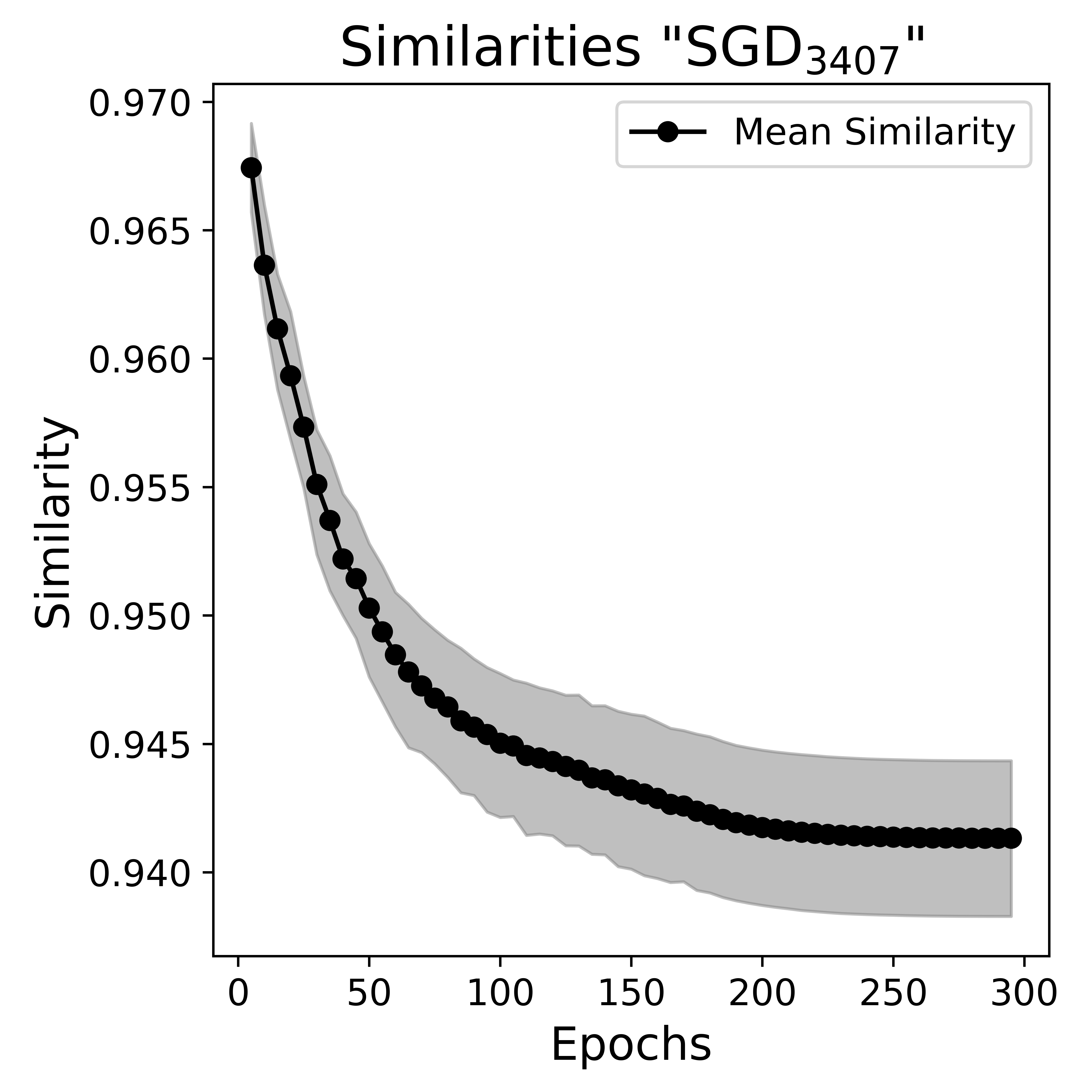}
\includegraphics[height=0.21\textwidth, trim={1cm 0cm 0cm 0cm}, clip]{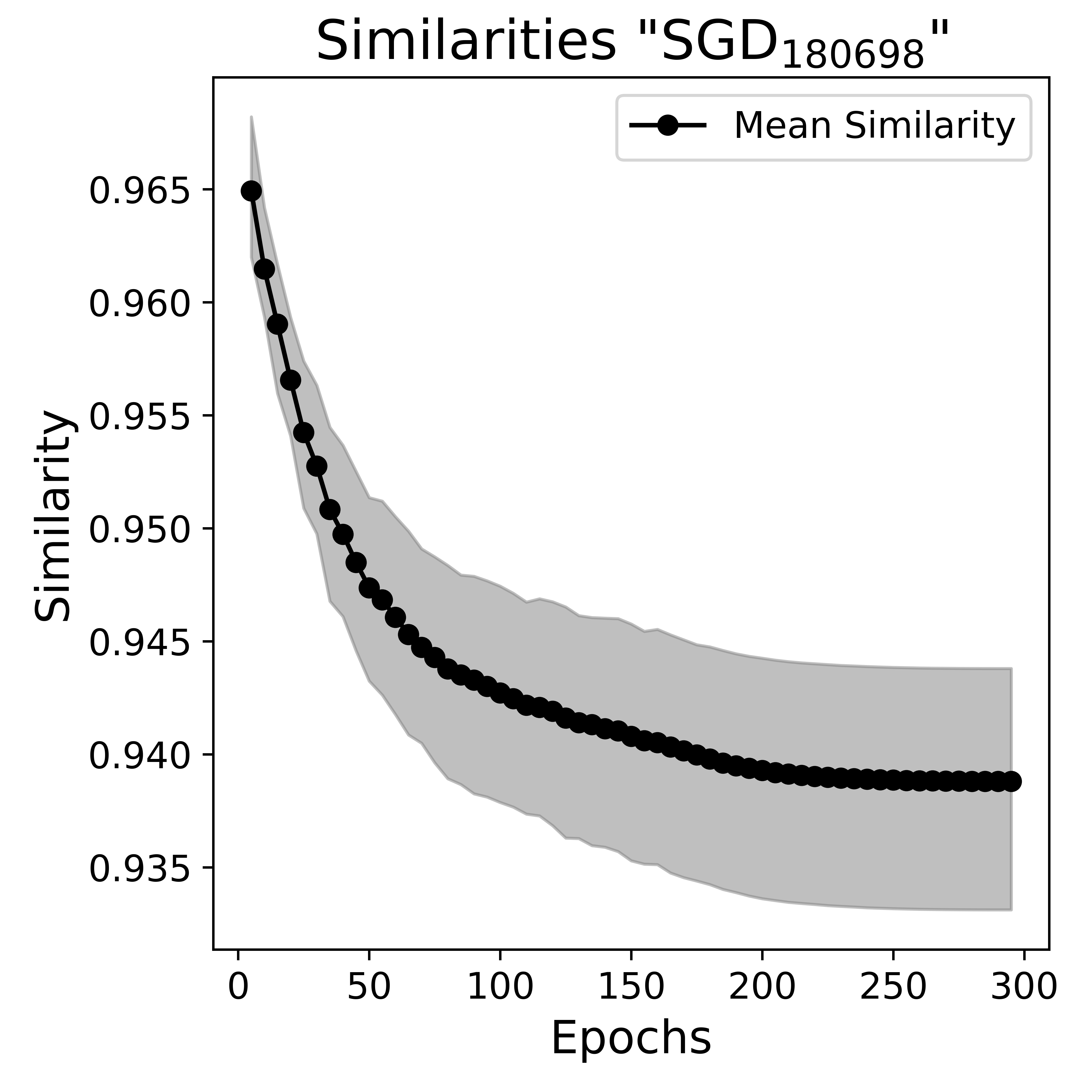}
\includegraphics[height=0.21\textwidth, trim={1cm 0cm 0cm 0cm}, clip]{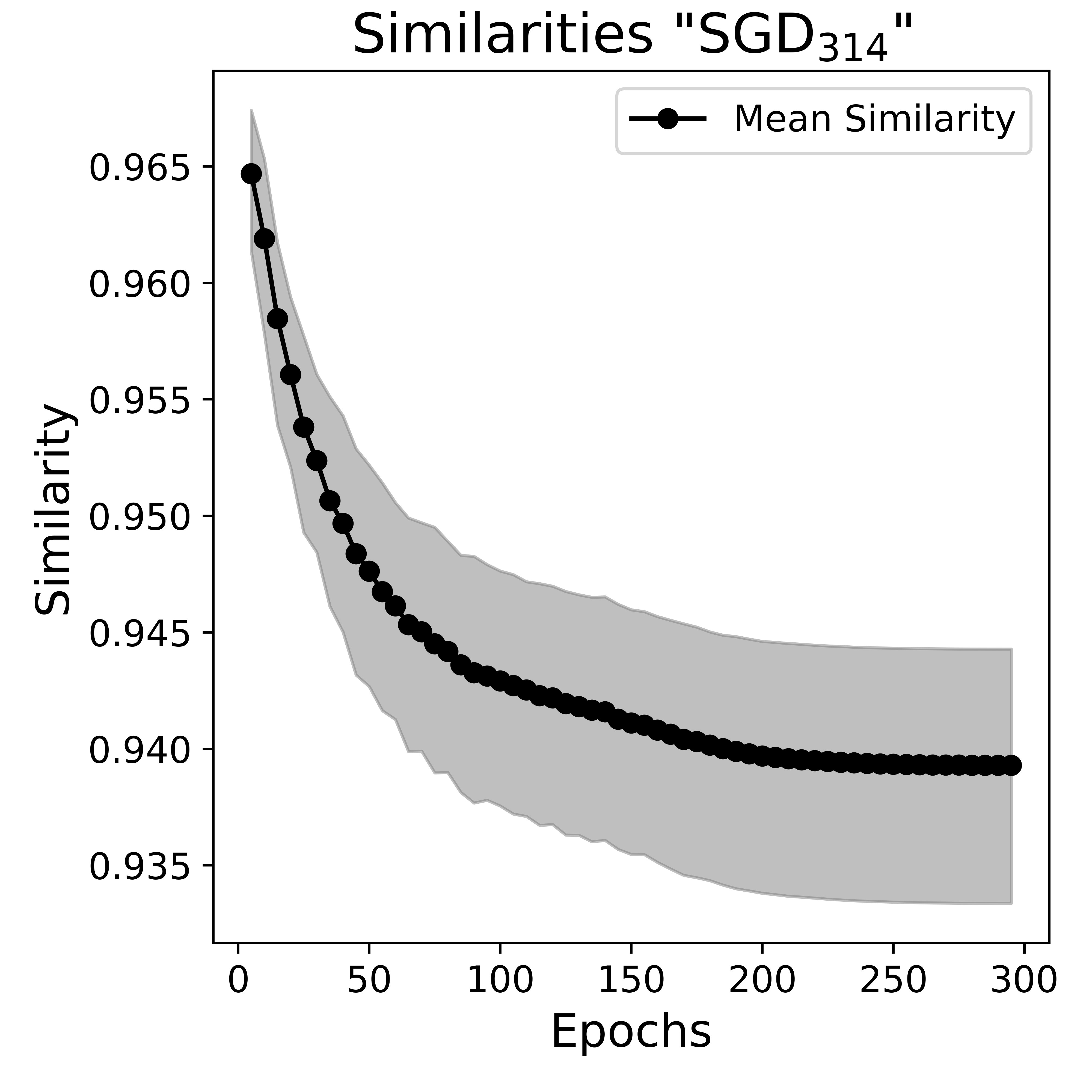}
\end{minipage}
\begin{minipage}{0.02\textwidth}
\rotatebox{90}{CBIS-DDSM}
\end{minipage}%
\begin{minipage}{0.97\textwidth}
 \includegraphics[height=0.215\textwidth]{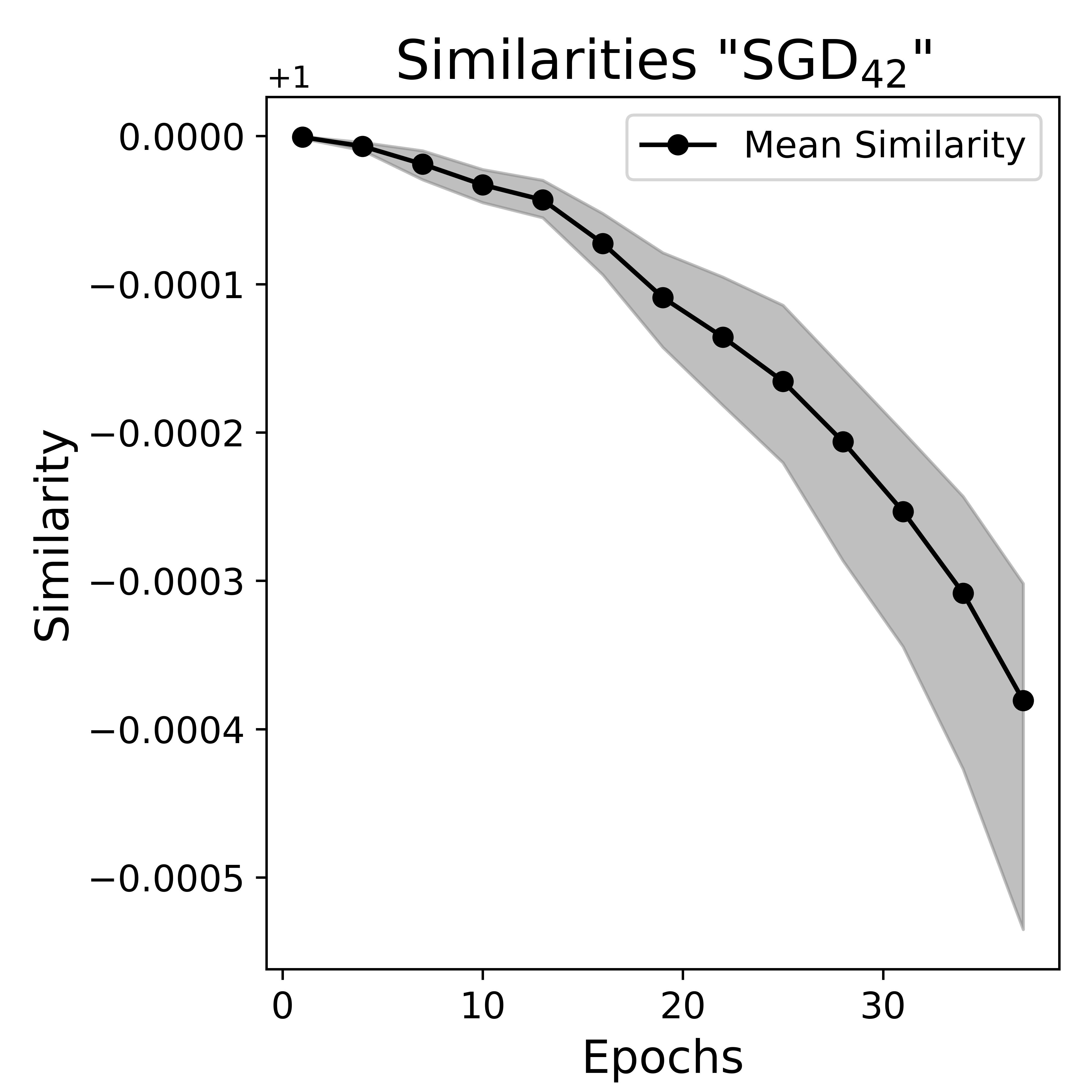}
\includegraphics[height=0.215\textwidth, trim={1.8cm 0cm 0cm 0cm}, clip]{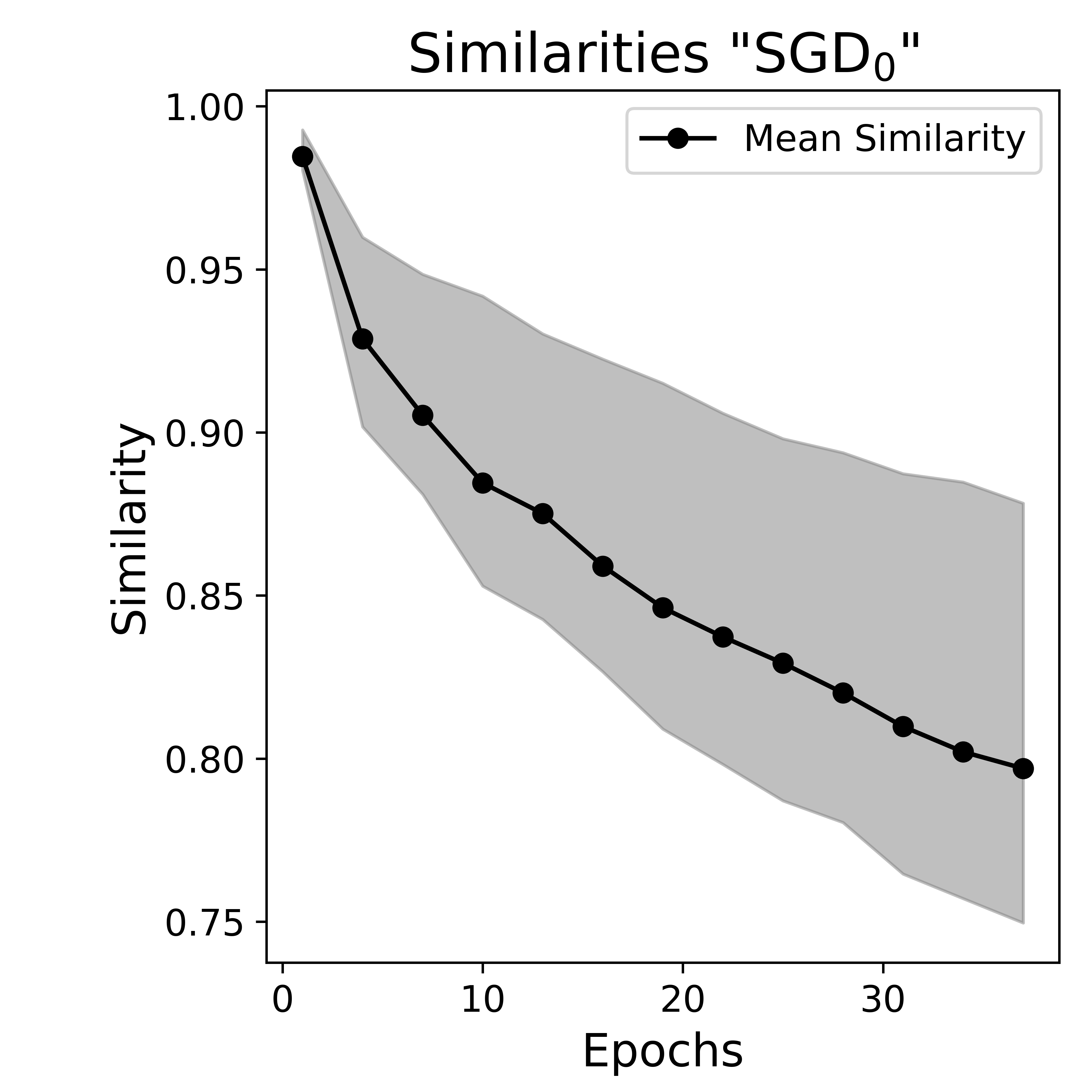}
\includegraphics[height=0.215\textwidth, trim={1.3cm 0cm 0cm 0cm}, clip]{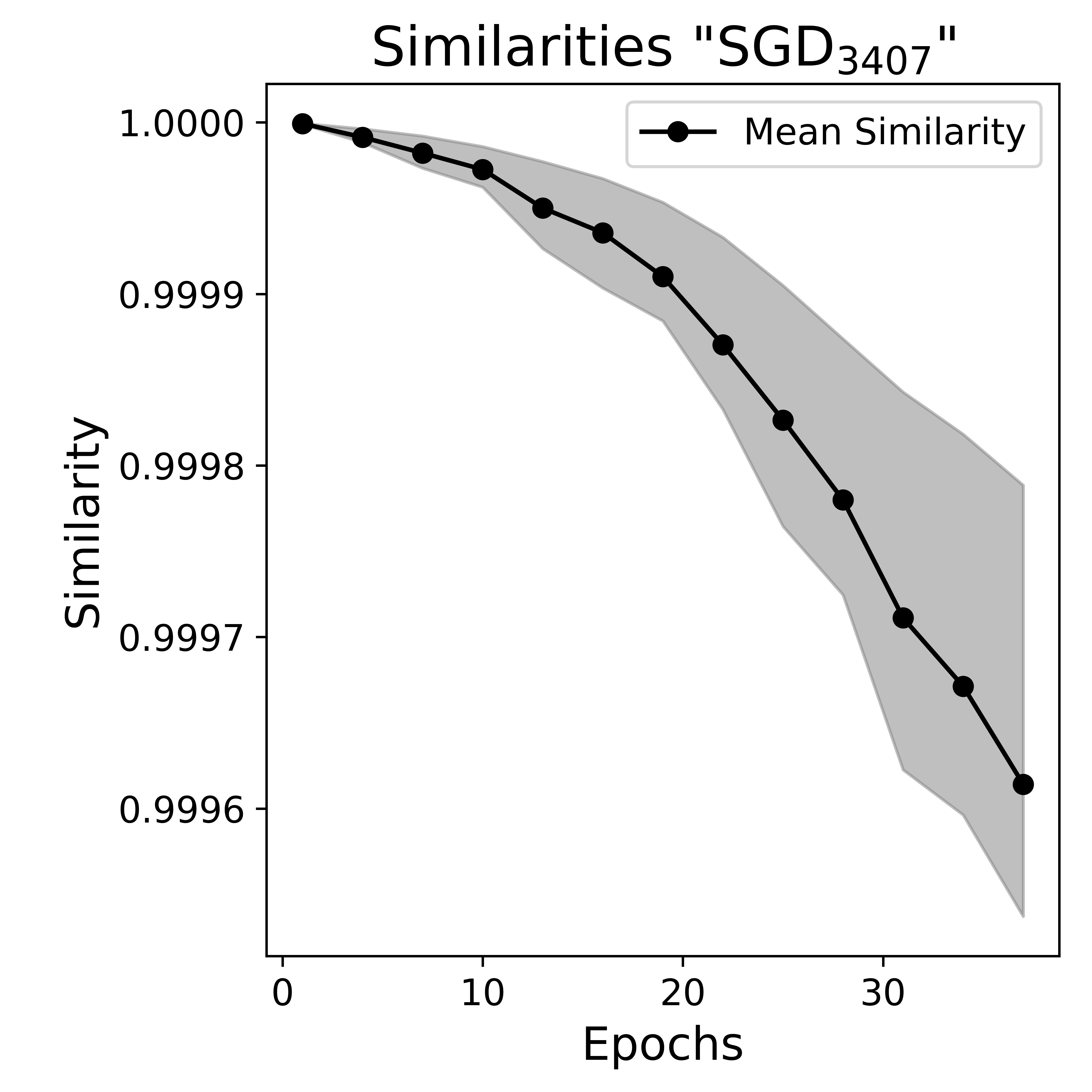}
\includegraphics[height=0.215\textwidth, trim={1.3cm 0cm 0cm 0cm}, clip]{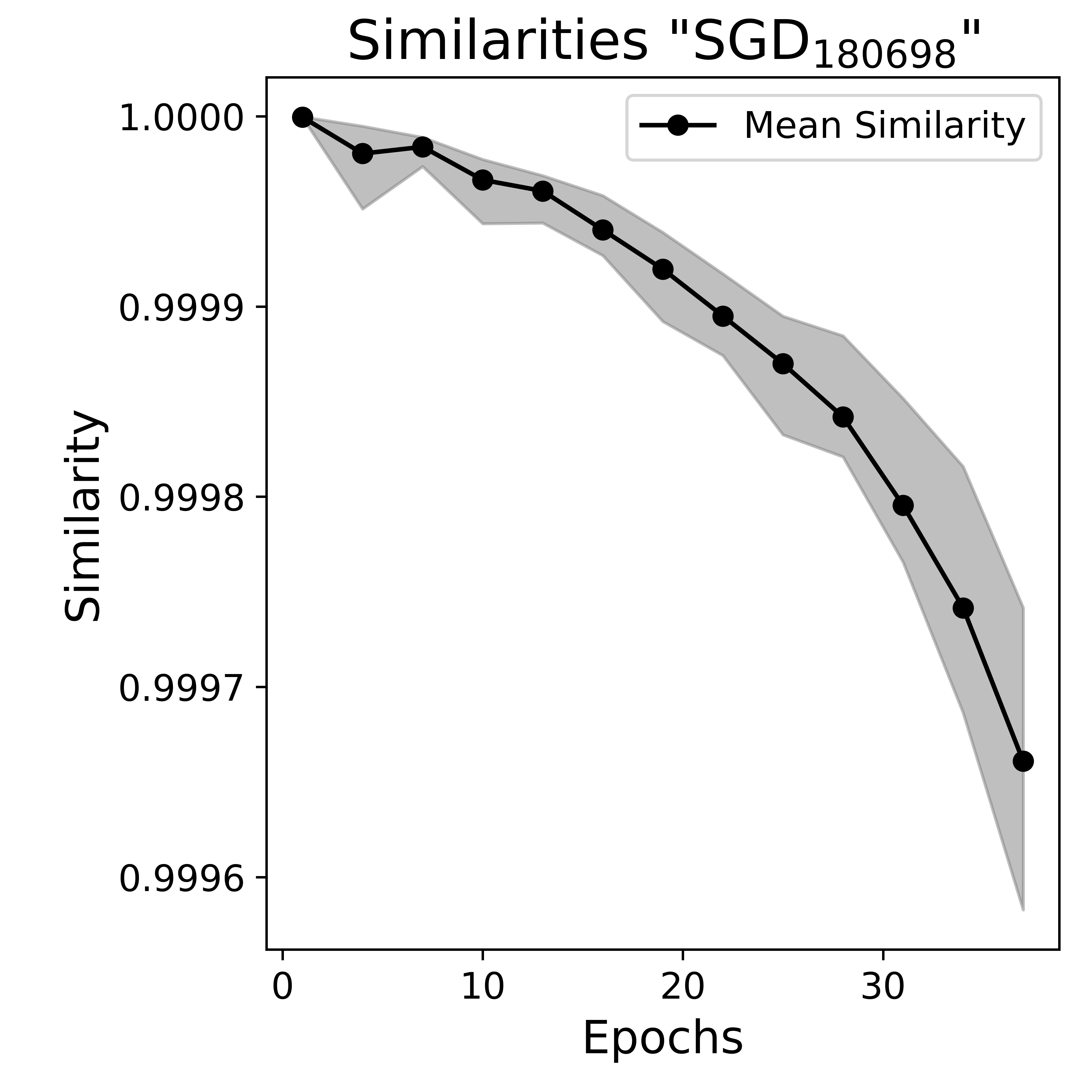}
\includegraphics[height=0.215\textwidth, trim={1.3cm 0cm 0cm 0cm}, clip]{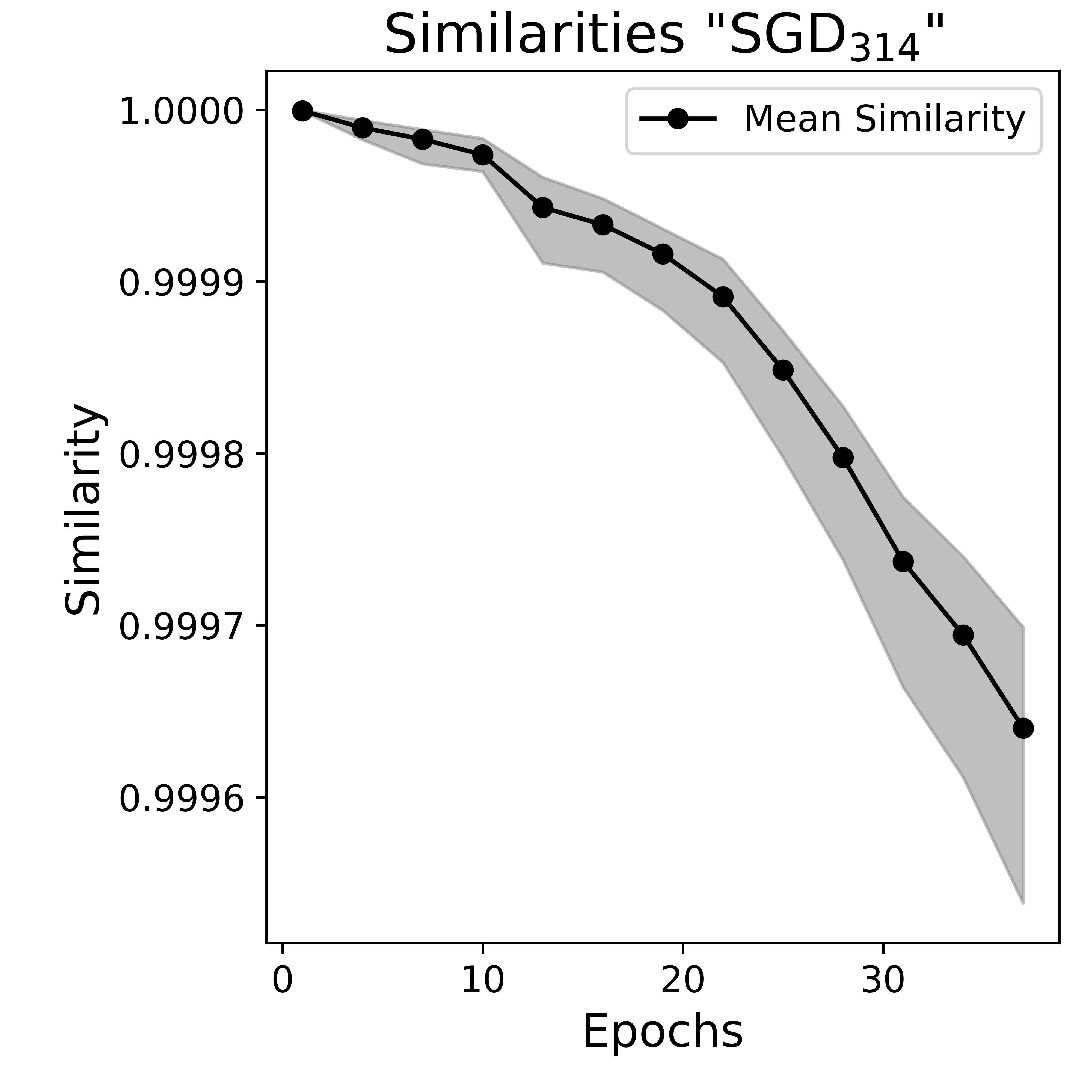}
\end{minipage}
\begin{minipage}{0.02\textwidth}
\rotatebox{90}{SDNET}
\end{minipage}%
\begin{minipage}{0.97\textwidth}
 \includegraphics[height=0.209\textwidth]{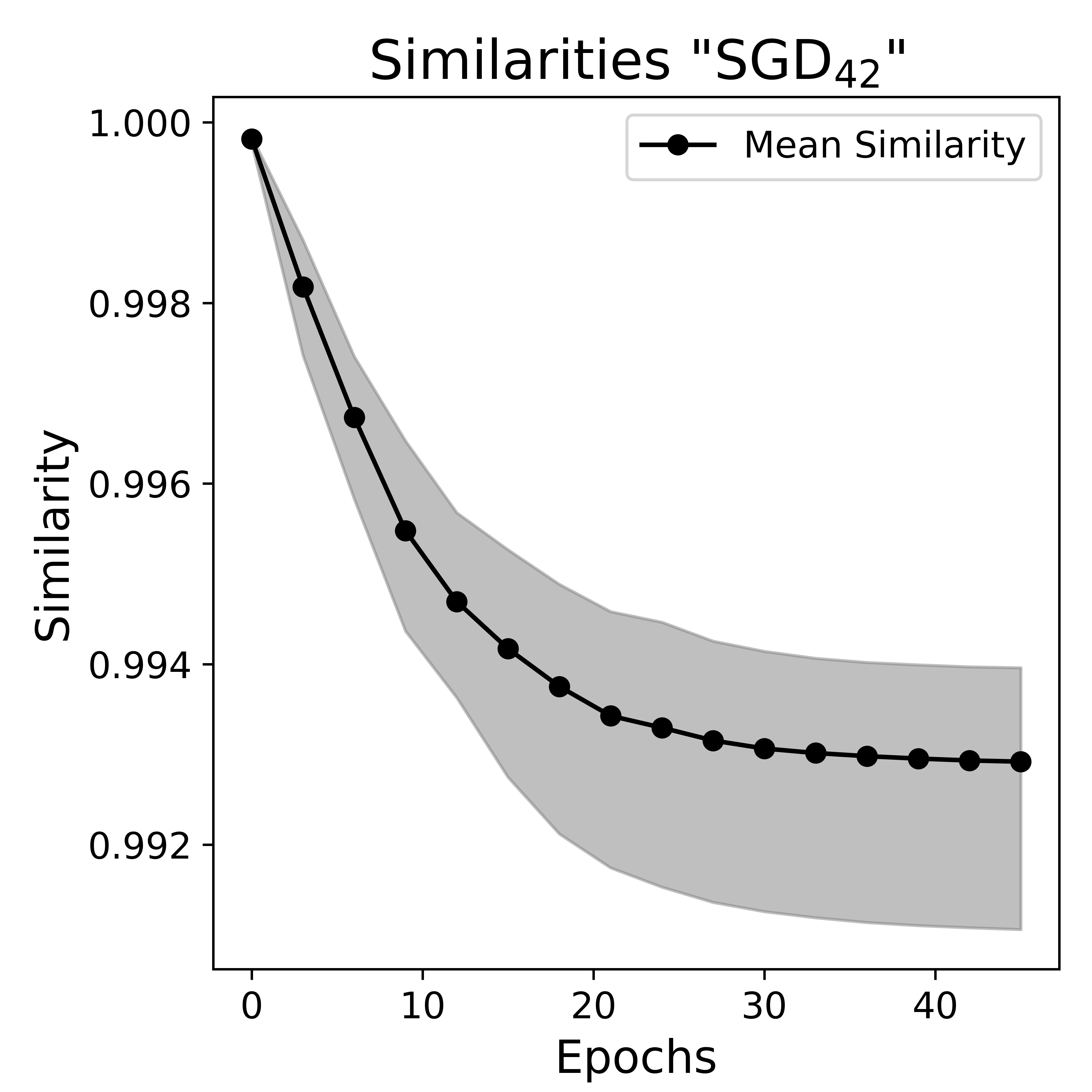}
\includegraphics[height=0.209\textwidth, trim={1cm 0cm 0cm 0cm}, clip]{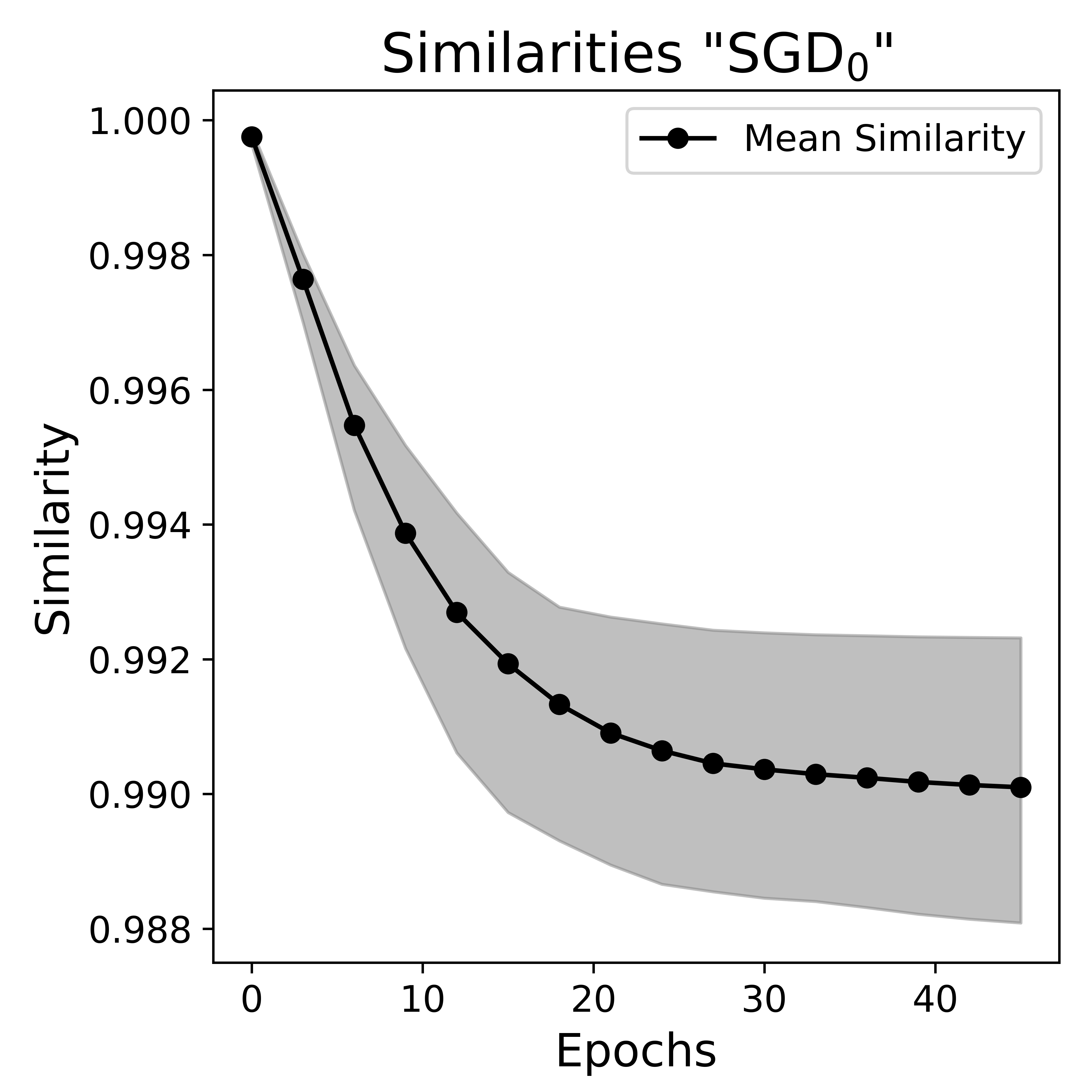}
\includegraphics[height=0.209\textwidth, trim={1cm 0cm 0cm 0cm}, clip]{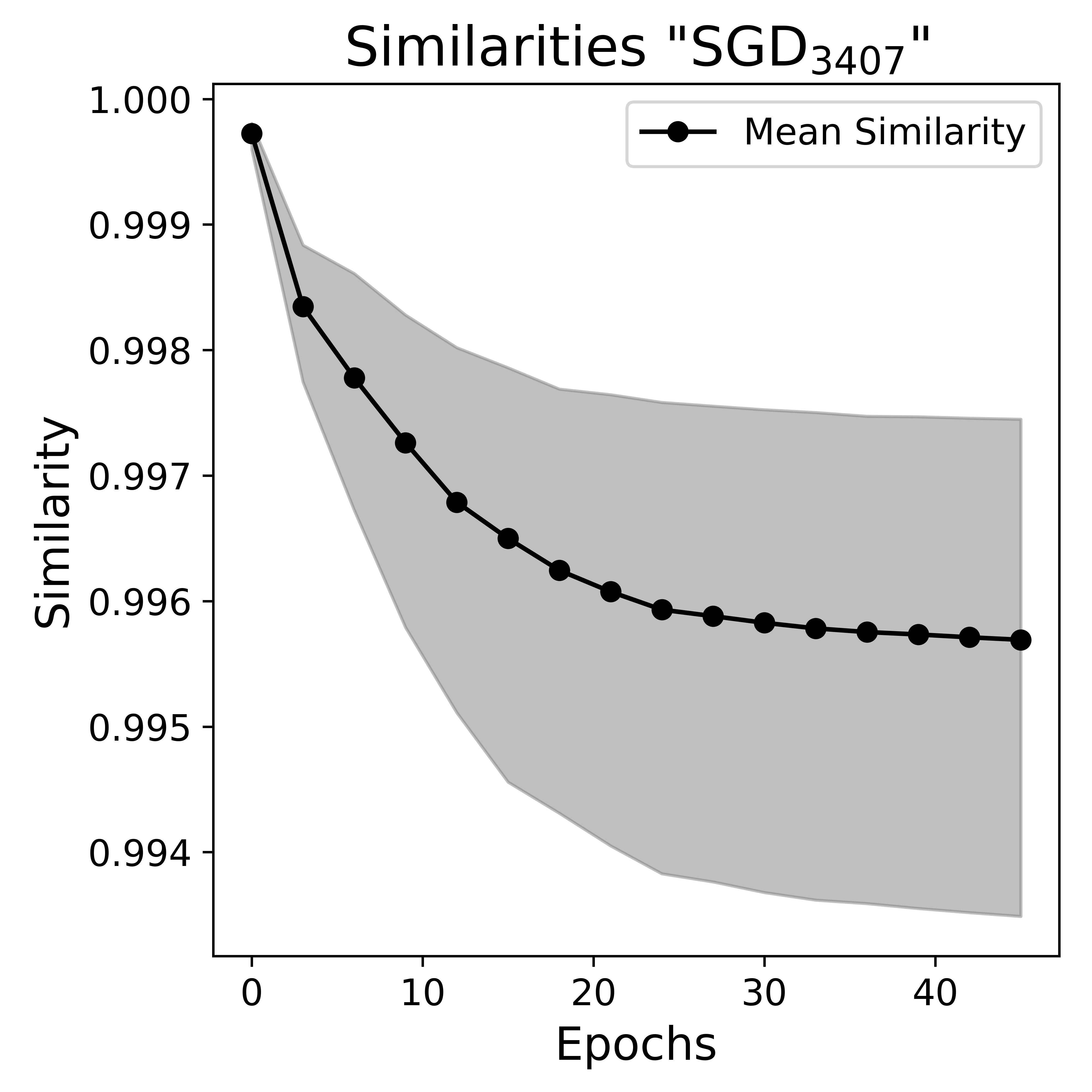}
\includegraphics[height=0.209\textwidth, trim={1cm 0cm 0cm 0cm}, clip]{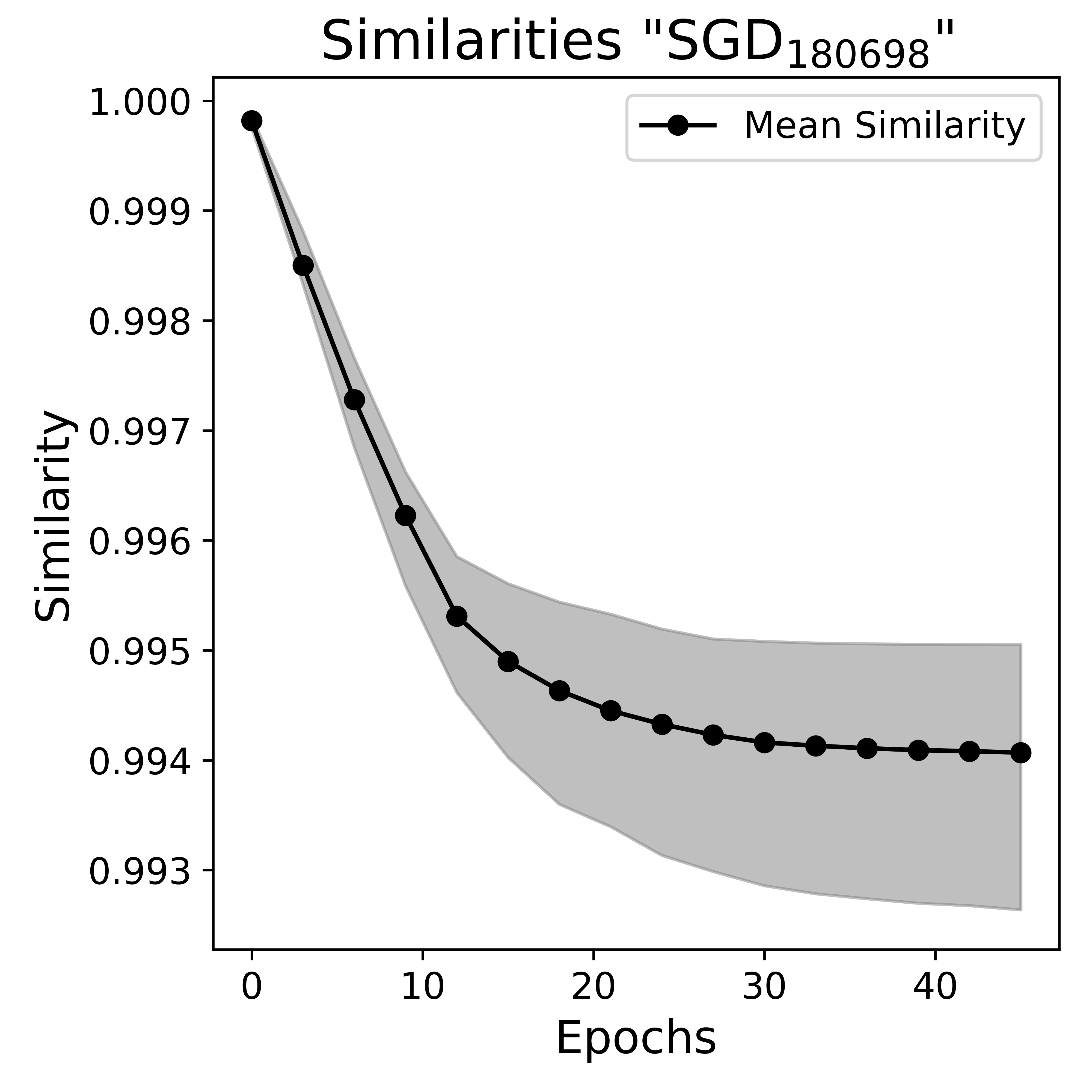}
\includegraphics[height=0.209\textwidth, trim={1cm 0cm 0cm 0cm}, clip]{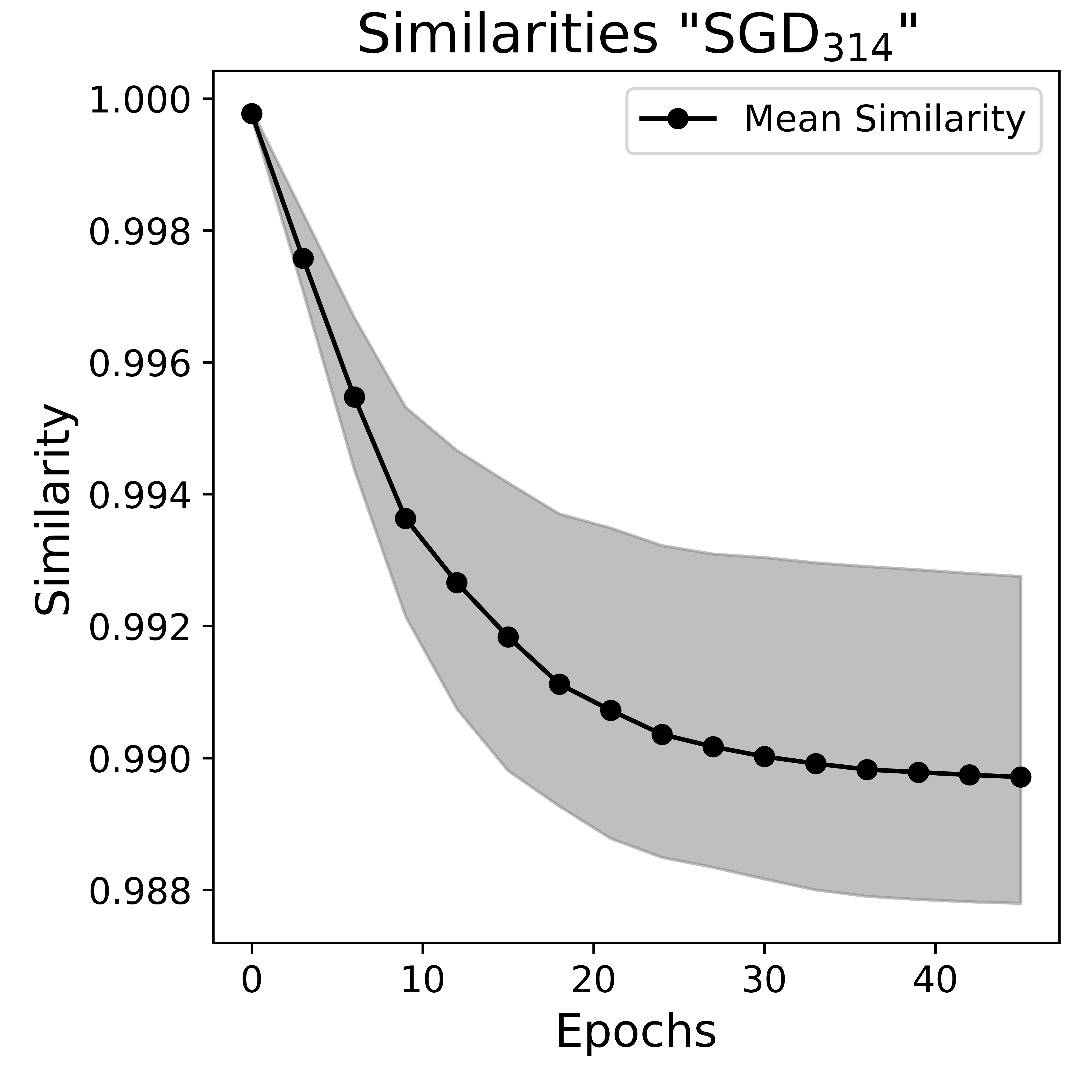}
\end{minipage}
  \caption{Cosine similarity of embeddings in the last linear layer of the models with respect to epoch number. Showing max, min, and mean cosine similarity values with SGD optimizer for each dataset (rows) and seed (column).}
  \label{fig:similarities_sgd}
\end{figure*}

Figs.~\ref{fig:similarities_adam} and~\ref{fig:similarities_sgd} show the  similarities of embeddings across datasets and optimizers. The mean similarities provide a general overview, while minimum similarities show the extreme points of two runs affected by inherent CUDA execution randomness. The minimum similarity provides insight into the maximum potential impact of this randomness. In the figures, a value of 1 indicates that the embeddings were identical, whereas a value of 0 indicates that they were completely orthogonal.

 Fig.~\ref{fig:similarities_adam} illustrates the last layer similarities for the ADAM optimizer across the three datasets. 
 For the CIFAR-10 dataset, we observe the lowest similarity and largest variations. In later epochs, there seems to be a convergence of embedding representations as similarity values stabilize.
 In the CBIS-DDSM dataset, there is a consistent decline in similarity. This consistent reduction may be attributed to the effects of fine-tuning from ImageNet~\cite{5206848}-initialized weights rather than training from scratch. For the SDNET dataset, a pronounced initial drop is observed, indicating a swift divergence of weights. Yet, the similarities in subsequent epochs decline more gradually, hinting at a plateau in weight divergence.

Fig.~\ref{fig:similarities_sgd} shows the embedding similarities for the SGD optimizer. 
Overall, we observe similar trends as for the ADAM optimizer, but less pronounced differences in similarities.
Again, CIFAR-10 has the lowest similarity and largest variation, but similarities are much higher as with the ADAM optimizer.
For the CBIS-DDSM dataset, there is a constant decrease in similarity, but embedding differences are subtle.
On the SDNET dataset, we again observe an initial drop, followed by a plateau. However, again, differences are subtle.

\label{src:05_environmental_impact}
\section{Environmental Impact of the Experiments}

In scientific research, it is crucial to consider not only the direct results of experiments but also the broader implications and consequences of the research process. While the following environmental assessment is not directly tied to our primary results, it represents an essential facet of our experiments. We believe it is our responsibility to report on the environmental footprint of our work, given the increasing global emphasis on sustainability and the environmental impact of computational practices. Furthermore, we posit that the environmental implications of computational experiments are becoming increasingly significant in the context of sustainable research practices. This perspective aligns with the findings of Ulmer et al.~\cite{ulmer-etal-2022-experimental}, emphasizing the importance of understanding and reporting the environmental consequences of experimental work.

Our experiments were conducted using HPC resources located in Essen, Germany. The region's electricity generation has a carbon efficiency of \(0.385 \, \text{kgCO}_2\text{eq/kWh}\)~\cite{ourworldindata2023}, with approximately \(43\%\)~\cite{statista2022energy} of the electricity being sourced from fossil fuels. To estimate the carbon footprint of our experiments, we utilized the Machine Learning Impact calculator, as presented by Lacoste et al.~\cite{DBLP:journals/corr/abs-1910-09700}. This calculator provides a comprehensive framework to quantify the carbon emissions associated with machine learning experiments, considering both the energy consumption of computational resources and the carbon efficiency of the electricity source.

\begin{table}[t]
  \centering
  \caption{Energy consumption and carbon dioxide (\(\text{CO}_2\)) emission for different tasks. The table shows the energy consumption in kilowatt-hours (kWh) and the corresponding \(\text{CO}_2\) emissions in kilograms (kg) under two categories: Experiment runs and All runs. The SUM row provides the total energy consumption for each run type, and the \(\text{CO}_2\) row calculates the emissions based on the formula \(\text{CO}_2\) (kg) = Energy (kWh) \(\times\) 0.385.}
  \label{tab:energy_consumption}
  \begin{adjustbox}{width=\columnwidth,center}
  \begin{tabular}{+l^c^c}
      \toprule\tabhead
      & Experiment runs & All runs \\
      \otoprule
      Cifar-10 (kWh) & 69.89 & 117.37 \\
      CBIS-DDSM (kWh) & 116.04 & 134.594 \\
      SDNET (kWh) & 89.29 & 114.78 \\
      SUM (kWh) & 275.22 & 366.744 \\
      \(\text{CO}_2\) (kg) & 105.06 & 141.69 \\
      \bottomrule
  \end{tabular}
  \end{adjustbox}
\end{table}

From Table \ref{tab:energy_consumption}, it is evident that while the energy consumption and associated carbon emissions for the reported experiments (``Experiment runs'') might not be significant, the overall environmental impact is considerably higher when accounting for all computational activities, including tests, debugging, and experimental setups (``All runs''). This highlights the broader environmental cost of the entire research process, not just the final reported results. It underscores the importance of energy-efficient algorithms and practices in machine learning research, especially in regions heavily reliant on fossil fuels for electricity generation.

\label{src:06_conclusion}
\section{Discussion and Conclusion}

From the analysis presented in Table~\ref{tab:comparison_adam_sgd_f1}, it becomes clear that CUDA-randomness significantly influences performance variability, with the impact varying across three datasets and two optimizers. The choice of optimizer shows dependency on the specific dataset, indicating that the interplay between optimizer and dataset characteristics is crucial. Moreover, we observe that different performance metrics exhibit varied levels of variability, suggesting that the choice of metric is pivotal in understanding the performance landscape. Additionally, variability introduced by different seed values is non-negligible, further complicating the performance analysis. Most notably, the mammography task demonstrates significantly greater performance variance compared to the other datasets, highlighting the task-specific nature of CUDA-randomness effects.

In examining the tradeoffs between deterministic and non-deterministic modes, our findings reveal nuanced differences in performance and runtime across various configurations (cf. Tables~\ref{tab:comparison_adam_sgd_f1} and~\ref{tab:runtime_comparison_adam_sgd}). While Zhuang et al.~\cite{zhuang2022randomness} and the PyTorch documentation\footnote{\url{https://pytorch.org/docs/stable/notes/randomness.html
} \\ Accessed: 2023-10-09} suggest that deterministic execution generally incurs higher runtime and lower performance, our experiments show this is not always the case. The performance differences between the two modes are usually within a 1\% margin for CIFAR-10 and SDNET datasets, indicating that deterministic settings could be favored for reproducibility without significantly sacrificing performance. However, for the mammography task, the performance gap occasionally exceeds 1\%, challenging the assumption that deterministic operations consistently yield higher performance or vice versa. Additionally, runtime analyses revealed that deterministic execution does not invariably lead to longer training times across all datasets. These findings emphasize the importance of considering specific dataset characteristics and algorithmic choices in PyTorch when evaluating the tradeoffs between deterministic and non-deterministic modes.

Our analysis of statistical significance in performance differences (cf. Table~\ref{tab:summary_observations}) showed that in particular the real-world mammography task is heavily influenced by CUDA randomness. We not only observed significant differences in all runs of the ADAM optimizer, but also the largest deviation of up to 1.68\% in all configurations that were significant (both for ADAM and SGD). This indicates that seed value selection can be crucial for the reproducibility of results in this task, and that the choice of optimizer may heavily influence the robustness of results.

To enhance reproducibility and robustness in deep learning research, we propose several strategies based on our study's insights. Firstly, adopting fully deterministic settings can offer significant benefits, as our results indicate, by zeroing variability and improving reproducibility. Conducting experiments across a range of seed values is also essential, allowing for a deeper understanding of how specific configurations impact outcomes. Moreover, ensuring full disclosure of all experimental settings, including configurations and software versions, is critical for enabling others to replicate and validate findings. 

Due to time and hardware constraints, runs per configuration were limited. Future studies could enhance robustness and depth of understanding by increasing the number of experimental runs. Our selected datasets are representative of application areas, but are still limited in scope. Future research could explore larger or more diverse datasets, such as ImageNet and CIFAR-100, to assess reproducibility across different scales and types of data. The selection of hyperparameters, informed by existing literature, suggests that a more exhaustive search might reveal further insights into hyperparameter influence on randomness. This study's focus on CUDA-induced randomness identifies a gap in the literature, indicating the necessity for additional research on computational consistency within CUDA architecture. 
Moreover, exploring the impact of different neural network architectures and distributed deep learning settings on reproducibility could be a further area of investigation. Lastly, considering different computational frameworks beyond PyTorch, such as TensorFlow, could offer a broader perspective on reproducibility challenges.

\noindent In summary, we derived the following insights from our study.
 
\textbf{Variability might be due to random seeds and chosen optimizers.} Empirical evidence highlights that model performance and variance are influenced by different seed settings and optimizer selections. Evaluating models across a range of seed configurations is essential for identifying setups that not only achieve optimal performance but also maintain minimal variance. 
Though widely used, the ADAM optimizer showed comparably lower robustness against CUDA-induced randomness in our study.

\textbf{Determinism introduces minor computational overhead and might slightly decrease predictive performance.} Our results show that deterministic execution usually introduces a negligible computational overhead while it is advantageous for reproducibility. Nonetheless, determinism can enhance performance by up to 2\% or result in reductions exceeding 1\%. This performance variability necessitates a careful decision-making process regarding the use of deterministic versus non-deterministic execution, based on the specific goals and performance criteria of the model.
    
\textbf{The effect of randomness is domain-specific.} The influence of randomness during training exhibits significant variation across different domains, with the medical imaging domain, exemplified by the mammography dataset, being particularly susceptible to randomness. This variation accentuates the reproducibility challenges in medical imaging and emphasizes the importance of domain-specific approaches in model training and evaluation.

\section*{Acknowledgments}
We thank the doctoral researchers from the Research Training Group 2535 Knowledge- and Data-Based Personalization of Medicine at the Point of Care (WisPerMed), Hendrik Damm, Helmut Becker, Tabea Pakull, and Mikel Bahn, who reviewed this paper and provided valuable comments.

\bibliography{bibliography}

\appendix
\label{src:07_appendix}
\section{Appendix}

\subsection{Reproducing the Results}

The datasets utilized in this study are publicly accessible. Details are as follows:

\begin{itemize}
    \item CIFAR-10 Dataset: The dataset can be accessed from the University of Toronto's website.\footnote{\url{https://www.cs.toronto.edu/~kriz/cifar.html}, last accessed 15 November 2023}
    \item SDNET2018 Dataset: It is available on the Utah State University's digital commons page.\footnote{\url{https://digitalcommons.usu.edu/all_datasets/48/}, last accessed 22 October 2023}
    \item CBIS-DDSM Dataset: The dataset is hosted on the Cancer Imaging Archive wiki.\footnote{\url{https://wiki.cancerimagingarchive.net/pages/viewpage.action?pageId=22516629}, last accessed 10 December 2023}
\end{itemize}

The codebase supporting this study is also open-source. The current GitHub repository is as follows:

\begin{center}
    \url{https://github.com/aix-group/repincv.git}
\end{center}

Within the repository, the code structure is organized into specific tasks:

\begin{itemize}
    \item task1: CIFAR-10 experiments.
    \item task2(SDNET 2018): Concrete Crack Detection experiments.
    \item task3(CBIS-DDSM): Breast Cancer Imaging experiments.
\end{itemize}

\noindent Always refer to the official repository as it can get fixes or updates in the future.\\

\subsubsection*{Instructions to Reproduce the Results}

While the CIFAR-10 and SDNET2018 datasets are set to auto-download if absent in the input directory, the CBIS-DDSM requires manual downloading due to its voluminous size. Once obtained, it should be relocated to the input directory. Subsequent image preprocessing steps are documented within the aforementioned repository.

For the experimental sections of this study, the (W\&B) platform was employed for result tracking. To replicate the findings, adhere to the following procedure:

\subsubsection*{Securing the (W\&B) API Key}
\begin{itemize}
    \item Initiate by visiting the Weights \& Biases official site.
    \item Proceed to sign in or establish an account if not already registered.
    \item Access your personal settings to retrieve your unique API key.
    \item Copy the API key for upcoming phases.
\end{itemize}
For a guide on securing the (W\&B) API key, consult the platform's official documentation.\footnote{\url{https://docs.wandb.ai/quickstart}}

\noindent\textrm{Setup Instructions:}

\begin{lstlisting}
git clone https://github.com/mcmi-group/repincv.git
cd repincv

wandb login

conda create --name repincv_env python=3.8
conda activate repincv_env

pip install -r requirements.txt

python run_experiments.py --task <task_name>
\end{lstlisting}

First, clone the repository and navigate into it. Next, log in to WandB; when prompted, paste the API key you obtained in the previous step. Then, create a new Conda environment named \texttt{repincv\_env} with Python~3.8 and activate it. Install the necessary packages using the \texttt{requirements.txt} file, which is located in the root directory of the repository and contains the versions of the important packages used in the experiments. Finally, run the experiment script with \texttt{python run\_experiments.py --task \textless{}task\_name\textgreater{}}, replacing \texttt{\textless{}task\_name\textgreater{}} with one of the following: \texttt{task1}, \texttt{task2}, or \texttt{task3}.

The cumulative training duration for all experiments was approximately 900 hours, equivalent to nearly five weeks of continuous computation on a single NVIDIA RTX A6000 GPU.\\\\\\\\\\\\\\\\\\\\\\\\\\\\\\

\label{appendix:additional_tables}
\subsection{Additional Tables and Graphs from the Experiment Data}

\begin{table}[ht!]
\centering
\caption{Deterministic and non-deterministic runtime comparison for the CIFAR dataset with two optimizers, ADAM and SGD, expressed in minutes. The first column lists all the seed configurations. $\overline{A}_{ND}$ represents the non-deterministic mean accuracy for each seed configuration. $\overline{A}_{D}$ represents the deterministic accuracy for each seed configuration. $\sigma$ indicates the runtime variation. $\mu$ represents the mean runtime for deterministic and non-deterministic scenarios.}
\label{tab:appendix_cifar}
\resizebox{\columnwidth}{!}{%
\begin{tabular}{ccccccc}
\toprule
Config & Optimizer & $\overline{A}_{ND}$ & $\overline{A}_{D}$ & $\sigma$ & $\mu_{ND}$ (min) &  $\mu_{D}$ (min) \\
\midrule
\multirow{2}{*}{0} & ADAM & 0.925 & 0.922 & 0.0018 & 110.94 & 135.82 \\
                   & SGD  & 0.948 & 0.950 & 0.0027 & 105.27 & 137.18 \\
\midrule
\multirow{2}{*}{180698} & ADAM & 0.930 & 0.922 & 0.0015 & 114.51 & 133.93 \\
                        & SGD  & 0.947 & 0.949 & 0.0016 & 105.44 & 135.97 \\
\midrule
\multirow{2}{*}{314} & ADAM & 0.923 & 0.922 & 0.0009 & 115.74 & 133.48 \\
                     & SGD  & 0.948 & 0.947 & 0.0043 & 103.90 & 134.58 \\
\midrule
\multirow{2}{*}{3407} & ADAM & 0.924 & 0.927 & 0.0010 & 114.07 & 135.18 \\
                      & SGD  & 0.947 & 0.946 & 0.0035 & 106.69 & 132.38 \\
\midrule
\multirow{2}{*}{42} & ADAM & 0.925 & 0.927 & 0.0030 & 112.38 & 135.30 \\
                    & SGD  & 0.947 & 0.948 & 0.0009 & 110.25 & 135.55 \\
\bottomrule
\end{tabular}
}
\end{table}

\begin{table}[ht!]
\centering
\caption{Deterministic and non-deterministic runtime comparison for the SDNET dataset with two optimizers, ADAM and SGD, expressed in minutes. The first column lists all the seed configurations. $\overline{F1}_{ND}$ represents the non-deterministic mean F1 Score for each seed configuration. $\overline{F1}_{D}$ represents the deterministic F1 Score for each seed configuration. $\sigma$ indicates the runtime variation. $\mu$ represents the mean runtime for deterministic and non-deterministic scenarios.}
\label{tab:appendix_sdnet}
\resizebox{\columnwidth}{!}{%
\begin{tabular}{ccccccc}
\toprule
Config & Optimizer & $\overline{F1}_{ND}$ & $\overline{F1}_{D}$ & $\sigma$ & $\mu_{ND}$ (min)  & $\mu_{D}$ (min) \\
\midrule
\multirow{2}{*}{314} & ADAM & 0.938 & 0.935 & 0.0018 & 166.63 & 153.20 \\
                     & SGD  & 0.929 & 0.922 & 0.0027 & 179.06 & 203.12 \\
\midrule
\multirow{2}{*}{180698} & ADAM & 0.940 & 0.939 & 0.0015 & 170.97 & 157.40 \\
                        & SGD  & 0.930 & 0.929 & 0.0016 & 164.38 & 159.08 \\
\midrule
\multirow{2}{*}{3407} & ADAM & 0.939 & 0.937 & 0.0009 & 176.83 & 213.62 \\
                      & SGD  & 0.926 & 0.928 & 0.0043 & 164.97 & 150.08 \\
\midrule
\multirow{2}{*}{0} & ADAM & 0.939 & 0.939 & 0.0010 & 161.15 & 161.33\\
                   & SGD  & 0.930 & 0.928 & 0.0035 & 210.25 & 221.62 \\
\midrule
\multirow{2}{*}{42} & ADAM & 0.936 & 0.941 & 0.0030 & 165.09 & 170.87 \\
                    & SGD  & 0.931 & 0.931 & 0.0009 & 157.04 & 152.25 \\
\bottomrule
\end{tabular}
}
\end{table}

\begin{figure*}[t]
      \centering
      
      \begin{subfigure}[b]{0.45\textwidth}
        \includegraphics[width=\textwidth]{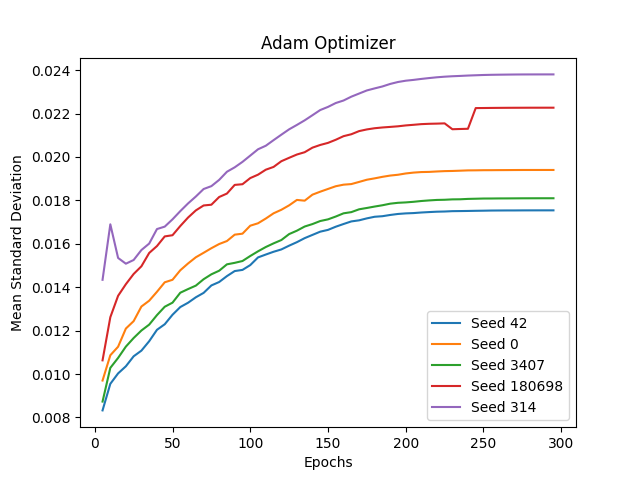}
        \caption{CIFAR-10 Variances for ADAM Optimizer}
        \label{fig:cifar10_adam_std}
      \end{subfigure}
      \hfill
      \begin{subfigure}[b]{0.45\textwidth}
        \includegraphics[width=\textwidth]{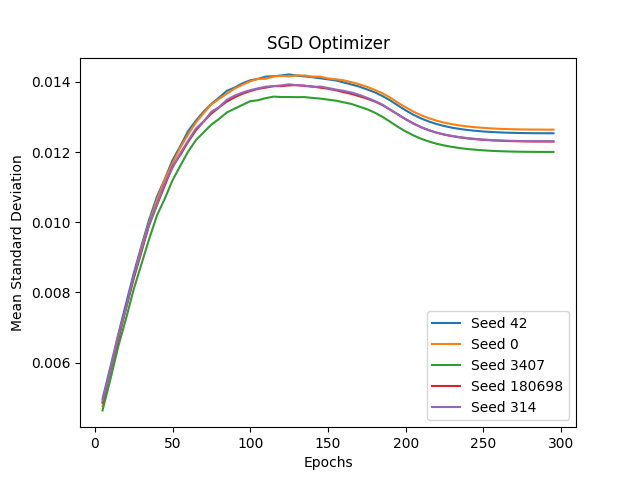}
        \caption{CIFAR-10 Variances for SGD Optimizer}
        \label{fig:cifar10_sgd_std}
      \end{subfigure}
      
      \begin{subfigure}[b]{0.45\textwidth}
        \includegraphics[width=\textwidth]{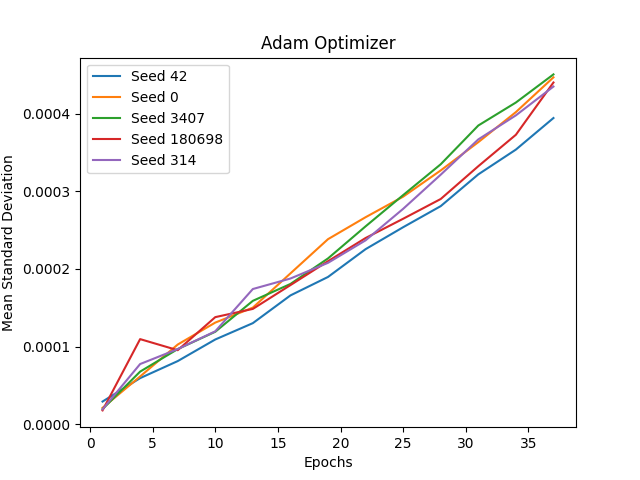}
        \caption{CBIS-DDSM Variances for ADAM Optimizer}
        \label{fig:cbisddsm_adam_std}
      \end{subfigure}
      \hfill
      \begin{subfigure}[b]{0.45\textwidth}
        \includegraphics[width=\textwidth]{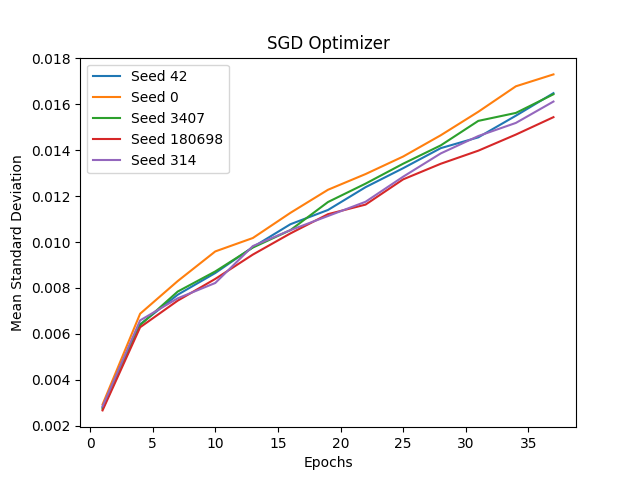}
        \caption{CBIS-DDSM Variances for SGD Optimizer}
        \label{fig:cbisddsm_sgd_std}
      \end{subfigure}
      
      \begin{subfigure}[b]{0.45\textwidth}
        \includegraphics[width=\textwidth]{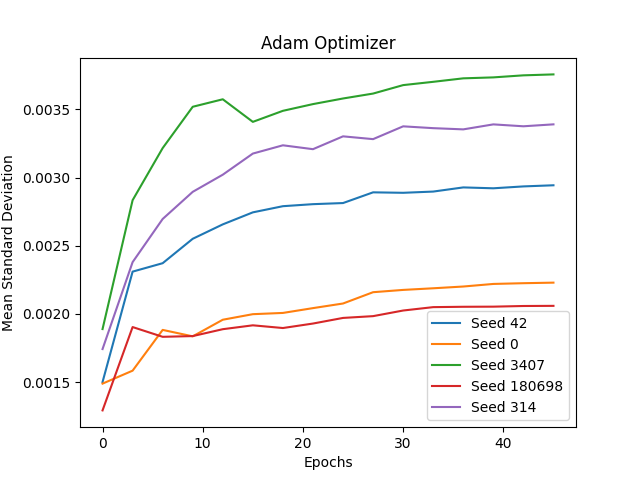}
        \caption{SDNET Variances for ADAM Optimizer}
        \label{fig:sdnet_adam_std}
      \end{subfigure}
      \hfill
      \begin{subfigure}[b]{0.45\textwidth}
        \includegraphics[width=\textwidth]{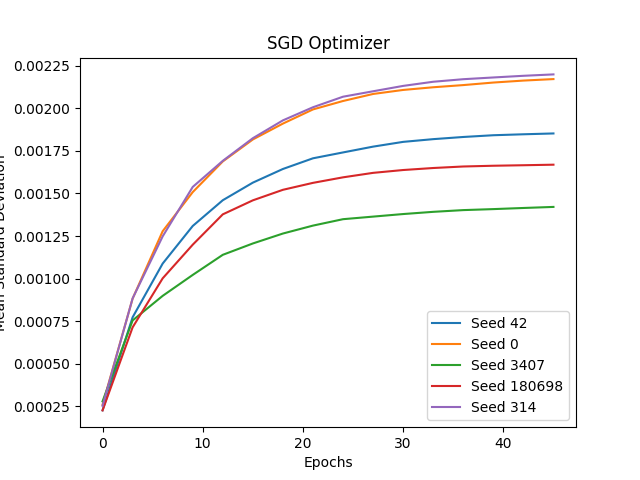}
        \caption{SDNET Variances for SGD Optimizer}
        \label{fig:sdnet_sgd_std}
      \end{subfigure}
      
      \caption{Mean standard deviation plots with respect to epoch number for each optimizer and dataset across five different seed configurations. The figures display the mean standard deviation in the classification head for each model and its progression over the training course. The comparison between the SGD and ADAM optimizers is highlighted for three datasets: CIFAR-10, CBIS-DDSM, and SDNET. The plots reveal the trends in standard deviations, showcasing how the randomness introduced by CUDA affects the consistency of model performance. Notably, the figures demonstrate how certain seed values contribute to greater fluctuations, with the ADAM optimizer often exhibiting more pronounced variances compared to SGD, particularly in the later epochs.}

      \label{fig:variance_plots_std}
\end{figure*}

\end{document}